\pdfoutput=1

\documentclass[11pt]{article}

\usepackage{acl}

\usepackage{times}
\usepackage{latexsym}

\usepackage[T1]{fontenc}

\usepackage[utf8]{inputenc}

\usepackage{microtype}

\usepackage{inconsolata}
\usepackage{url}
\usepackage{linguex}
\usepackage{babel}
\usepackage{natbib}
\usepackage[useregional]{datetime2}
\usepackage{graphicx} 
\usepackage{amsmath}
\usepackage{tcolorbox}
\usepackage{amsfonts}
\usepackage{amssymb}
 \usepackage{mathtools} 
\usepackage{caption}
\usepackage{subcaption}
\usepackage{dsfont}
\usepackage{changepage}
\usepackage{bbm}
\usepackage{bm}
\usepackage{wrapfig}
\usepackage{multirow}
\usepackage{multicol}
\usepackage{natbib}
\usepackage{tikz}
\usepackage{amssymb,amsmath,linguex,amsthm}
\usepackage{amsmath}
\usepackage{algorithm}
\usepackage[noend]{algpseudocode}

\usepackage{tcolorbox}

\theoremstyle{definition}
\newtheorem{definition}{Definition}

\newtheorem{prop}{Proposition}

\newcommand{\hidden}[1]{}

\newcommand{\fh}{\hat{f}}

\let\emptyset\varnothing

\title{Strong hallucinations from negation and how to fix them}
\author{Nicholas Asher \\
  CNRS,  IRIT  \\
  118 route de Narbonne  \\
  Toulouse, France \\
  \texttt{asher@irit.fr} \\\And
  Swarnadeep Bhar\\
  IRIT / Université Paul Sabatier\\
 118 route de Narbonne \\
  Toulouse, France  \\
  \texttt{swarnadeep.bhar@irit.fr} \\}
\date{\today}

\begin{document}
\maketitle

\begin{abstract}
Despite great performance on many tasks, language models (LMs) still struggle with reasoning, sometimes providing responses that cannot possibly be true because they stem from logical incoherence.  We call such responses \textit{strong hallucinations} and prove that they follow from an LM's computation of its internal representations for logical operators and outputs from those representations.  Focusing on negation, we provide a novel solution in which negation is treated not as another element of a latent representation, but as \textit{an operation over an LM's latent representations that constrains how they may evolve}. We show that our approach improves model performance in cloze prompting and natural language inference tasks with negation without requiring training on sparse negative data. 

\end{abstract}

\section{Introduction}

This paper investigates and addresses reasoning problems for language models (LMs) that stem from their representation of the meaning of logical terms.  Underlying such problems, which we call {\em strong hallucinations}, are faulty representations that lead LMs into logical incoherence, and hence into error, regardless of the facts.   Concentrating on negation, we 
prove that how an LM computes meaning representations for negation and how that affects outputs inevitably lead to logical error. 

We argue that we can eliminate these errors and strong hallucinations by altering the way an LM treats and hence interprets logical operators.  We illustrate our approach with negation.  
Our idea is simple but radical: whereas for LMs all tokens in a context play the same role in determining an attention based probability distribution over next tokens or ``continuations'' of the context, we claim logical operators should operate differently from non logical expressions.  For instance, negation does not contribute just another token to a latent representation via linear algebraic operations; it provides constraints that shift an input probability distribution $\Pi$ that is determined by the tokens $\phi$ in its scope to a distribution $\Pi'$, with $\Pi'(\neg \phi) = 1 - \Pi(\phi)$.  
These constraints, along with others for sequences or conjunction, allow us to calculate algorithmically and recursively transitions from input distributions over only positive content to appropriate distributions for a continuation containing logical operators.  
The output distribution not only faithfully reflects the logical meaning of negation but also other first order logical operators. 
 
 While hallucinations are often observed for text generation, these settings are difficult to evaluate.  Thus, we evaluate our algorithmic or hybrid approach, we call it $\Lambda$ for negation, on three easily evaluable tasks: yes|no question answering, masked knowledge retrieval (MKR) and natural language inference (NLI) with negation, using both encoder LMs and LLMs. Nevertheless, we will illustrate $\Lambda$ on text generation in NLI.


$\Lambda$ yields good performance on MKR and NLI; we increase accuracy by 13\% on one NLI dataset and by 91\% on another.  Our datasets feature negation and are modified, linguistically validated versions of \cite{dagan:etal:2005,hossain-etal-2020-analysis}. $\Lambda$ uses an LM fine-tuned only on positive data, which avoids the problem of finding sparse negative data.  Our data sets code and the Llama3-8b model fine tuned on positive SNLI data are in the repository \url{https://github.com/swarnadeep8597/Negation.git}, and the model is on \url{https://huggingface.co/swarnadeep/snli-5-adapter/tree/main}

Section \ref{sec:background} provides background on hallucinations, distributions and linguistic meaning.  
Section \ref{sec:hallucinationsLM} proves strong hallucinations follow from basic assumptions about an LM's treatment of negation.  
Section \ref{sec:dynamic} details our positive proposal, and 
Section \ref{sec:empirical} provides experiments with our approach on Q\&A, MKR and NLI. Section \ref{sec:related} discusses related work. 

\section{Background} \label{sec:background}
{\bf LMs.} We assume LMs  trained on transformer architectures \cite{vaswani2017attention}  with multiple layers $i$ over very large corpora using masked or causal language modeling.  In decoder models, the decoder is trained to predict next token  $y_i$ given a context matrix $c_i$ and previously predicted tokens \{$y_1 , · · · , y_{i-1}$\} using a conditional distribution $\mu(y_i/y_1, ..., y_{i-1}, c_i)$. In encoder models, $c_i$ is typically the input to an MLP layer to make predictions.  $c_i$ is the LM's internal representation at $i$ of all the input tokens in its context and is a concatenation 
of multiple hidden states $z^h_j$, one for each attention head such that:
\begin{equation}
z^h_j = \Sigma^K_{k=1}\alpha^h_{j,k}W^h_Vx_k  \label{attn}
    \end{equation}
$\alpha_{j,k}^h$ are the attention weights providing the importance of token (representation) $x_k$ to $x_j$ and $W^h_V$ is the value weight matrix of head $h$. All tokens play a similar 
role 
given Equation \ref{attn}.\footnote{For motivation, see \cite{baroni:zamparelli:2010}; these representations contain rich linguistic information \cite{bert, roberta:2019,tenney:etal:2018,talmor:etal:2020,zhu:rudizicz:2020}.}

 This representation enables LMs to learn a function $\sigma: (C \times X)\rightarrow [0,1]$ that assigns probabilities for string $x \in  X \subset V^m$ given context $C \subseteq V^n$, where $V^n$ is the set strings of length $n$ with vocabulary $V$. For encoder and decoder models and tasks we consider here, $m$ = 1 typically.  So we will abstract away from architecture and learning details for our models (the tasks we examine all fit in this category); they learn a similar function from contexts to tokens of different kinds, though different loss functions may be used. 
 Generative LMs use conditionalization as in Equation \ref{calculation} to assign probabilities to novel strings $s = (w_1, w_2, ... , w_{m+1})$ given context $C$:  
\begin{equation} \label{calculation} \mu(s | C ) = \mu(w_1 | C) \times \mu(w_2 \ | C,w_1) \times \end{equation}
\vspace{0.02in}
\hspace*{0.9in}$... \times \mu(w_{m+1} | C,w_{1},...,w_m)$

Generative LMs and encoders use this distribution over strings to select ouputs or continuations given a context $C$.  Various methods ${\cal M}$ (maxarg, greedy, beam search, or sampling) over the distribution $\mu(. |C)$ have been proposed to find the maximally plausible continuation $\phi \in X^n$ of $C$.  An LM's affirmations, which we use in defining hallucinations below, are directly linked to this distribution.
\begin{definition}\label{affirm} $\fh$ affirms $\phi$ given $C$, if $\phi  = {\cal M}\{x \in X : \mu(x|C)\}$
\end{definition} 

 {\bf Strong Hallucinations.} In this paper, we examine a special kind of hallucination we call {\em strong hallucinations}.  To clarify strong hallucinations, we first provide a gloss for {\em hallucinations}.  Hallucinations involve an attitude like belief or affirmation towards a proposition or sentential content $p$.  An agent $A$'s degree of belief in $p$, $\mu(p)$, which is dependent on $A$'s evidence $E$, should determine their sincere assertion of $p$.  This holds for LMs too; they use probability distributions to determine their ``affirmations'' (output).  Now $A$'s degree of belief in $p$ may or may not correlate with a ``reasonable'' degree of belief in $p$, $\pi(p)$---a degree of belief that one expects a rational person to have.  On one type of hallucination, then, an agent $A$ with grounds or context $E$ hallucinates that $p$ if $\mu_A(p|E)$ diverges significantly from $\pi(p |E)$. 

 A strong hallucination by $A$ involves a reasoning mistake given evidence.  For an LM this evidence is a context or prompt.  The result of a strong hallucination is that $\mu$ leads $A$ to affirm a claim or set of claims that strongly diverge from any possible reasonable degree of belief given the evidence.  A special case is when the set of claims together with the evidence is logically unsatisfiable or cannot possibly be true together.  

LLMs have strong hallucinations.  Consider, for example this session  from LLama2-7b on the NLI task we detail in Section \ref{sec:empirical}. The task was is to decide whether a hypothesis H is entailed by a premise P. To help it with this task, we instructed Llama in a prompt to answer two questions: Q1 {\em can H be true if P is true?} and Q2 {\em can H be false if P is true?}. We instructed it to answer “entailment”(E) if it answered no to Q2 -- and “neutral” (not E) if it answered yes to Q1 and to Q2. Here is an example of the LLM's response (R):

\noindent
(R) {\em ``[INST]PREMISE: A young girl rides her bike HYPOTHESIS: This little girl is riding her bike Relation:[/INST] Based on the rules and procedure provided, I will ask myself 1) Can the hypothesis be true if the premise is true and 2) Can the hypothesis be false if the premise is true? Since the premise and hypothesis are the same, it is possible for the hypothesis to be true if the premise is true. Therefore, the answer to question1 is yes. Since the premise and hypothesis are the same, it is not possible for the hypothesis to be false if the premise is true. Therefore, the answer to question 2 is no. Based on my answers to the questions, I will predict a neutral relation between the premise and hypothesis” (not E).}

We assume that Llama has taken the context as given with the instructions in which we have if yes to Q1 and no to Q2, then it should infer entailment.  Llama answers correctly yes to Q1 and no to Q2, but apparently converts its "no" to Q2 into a yes answer or simply affirms not E despite accepting {\em A, B and if A and B, then infer E}. This is an example of a logical error that infects not only the LLM's reasoning but also, as in this case, its explanations of its behavior.  

We now provide a technical definition of strong hallucinations using probability distributions.  
We idealize rational distributions by considering {\em objective} distributions that capture the probabilities of events that are coherent with respect to probability axioms. $P$ represents the objective distribution over events or propositions described by strings in $X$ in the actual state, while  $P_w$ is an objective distribution for a logically possible way $w$ the world could be.  Let $\mu_{\fh}( . | E)$ be the conditional probability distribution of an LM $\fh$ over a set of truth evaluable strings $X$ that express a set of propositions that includes $p$.
\begin{definition} \label{divergence}
 $\fh$ {\em strongly hallucinates} that $p$, if $\mu_{\fh}(p | E)$ significantly diverges from $P_w(p | E)$ for any possible objective distribution $P_w$ and state $w$.\footnote{In the same spirit, we can define standard hallucinations: $\fh$ {\em hallucinates} that $p$, if $\mu_{\fh}(p | E)$ is significantly different from $P(p |E)$ and $\mu_{\fh}(E)$ significantly diverges from $P(E)$.}
\end{definition}

{\bf Probability and linguistic meaning} 
For epistemologists an ideal agent's  probability distribution $\kappa$ governing assertions should equal some objective distribution $P$ \cite{lewis:1981,hall:1994}.  
An ideal distribution $\kappa$ thus differs in an important but subtle way from LM distributions. $\kappa$ is {\em not} a probability distribution over sentences or strings but over propositions, what sentences express. To link with assertions, we use $\kappa$ to assign values to strings as follows: for any proposition $p$, if strings $\phi, \psi$ both express $p$, then $\kappa(\psi) = \kappa(\phi).$
In addition, the meanings of logical operators like $\neg$ (not), $\vee$ (or), $\wedge$  (and) impose the structure of a Boolean algebra $\mathbb{P}$ on the set of propositions.  
Hence, any probability function $\kappa$ respecting $\mathbb{P}$ that assigns values to sentences of a language verifies all the probability axioms (see Proposition 2 in Appendix A.1) and assigns every sentence expressing a logical truth probability 1, every inconsistent string probability 0 and if $\phi$ is a logical consequence of $\psi$ then $\kappa(\phi |\psi) = 1$.  
Conversely, each linguistic model for meaning defines such a probability function $\kappa$
(See Propositions 3, 4 Appendix A.1).  
Thus, an agent whose assertions are governed by $\kappa$, will avoid strong hallucinations.






\section{LMs, negation and strong hallucinations} \label{sec:hallucinationsLM}

Section \ref{sec:background} showed a tight connection between ideal distributions, meanings of strings that have a negation operator and conjunction or sequencing operator and probability axioms.  This connection does not hold for an LM's objective function that is a distribution over strings.

\begin{prop} \label{dutch}
Every LM $\fh$ whose 
outputs are governed by $\mu_{\fh}$ and Definition \ref{affirm} must strongly hallucinate if either:
(i)  $\sum_{x \in V^n \setminus \{\phi, \neg \phi\}} \mu_{\fh}(x)  > 0$;
(ii) $\mu_{\fh}$ assigns values to strings in $V^n$ that are logical truths or deductively valid reasoning steps.
\end{prop}
Proof: (i). Given the assumptions in (i), $\mu_{\fh}(\phi) + \mu_{\fh}(\neg \phi) < 1$.  But for every objective distribution $P$, $P(\phi) + P(\neg \phi) = 1$.  So $\mu_{\fh}$ diverges from any possible objective distribution. (ii). Any objective distribution must assign probability 1 to every logically true sentence $\ell$ of $V^*$, the set of all finite strings in vocabulary $V$.  
For each positive clause $Q$ there is such an $\ell$ of the form {\em Not ($Q$ and Not $Q$)}.  As a distribution over individual strings, $\mu$ cannot assign 1 to all such $\ell$.  Moreover, suppose given context the model offers a chain of deductive reasoning, $\phi. \ell. \ell'$.  But then by the axioms of probability and Equation 2, an objective distribution gives $P(\ell|\phi).P(\phi) = P(\phi)$.  So $P(\ell|\phi) = 1$. If $\mu(\ell|\phi) = 1$, however, $\mu$ assigns no other continuation of $\phi$ non $0$ probability; e.g., $\mu(\ell'.\ell |\phi) = 0$. This shows that LM distributions cannot respect the property of objective distributions that logically equivalent formulas receive the same probability.   In either case, by Definition 2, the LM strongly hallucinates.$\Box$
\noindent We note that a similar argument to that in the proof of Proposition \ref{dutch} (ii) shows that there are pairs of logically equivalent expressions to which $\mu_{\fh}$ will assign distinct values.

\cite{ramsey:1931,definetti:1937} linked the probability distributions underlying strong hallucinations to irrational behavior on bet. Our results correctly predict that LMs will place pathological bets (See Appendix G for an example). While most NLP practitioners won't care that their LM bets badly, reasoning errors like that in the example given in Section 1 are serious.  Suppose an LM must output a reasoning chain $R = (\phi, \ell_1, ...,\ell_r)$ in which $\ell_i$ in $R$ should follow deductively from previous elements in $R$.  
\begin{prop} \label{diverge}  If an LM $\fh$ 
outputs a reasoning chain $R$ with length $r$, $\mu_{\fh}$ will increasingly diverge from any possible objective distribution and $\fh$ will eventually surely hallucinate as $r$ increases. 
\end{prop}
\noindent
Consider a reasoning chain $R$ and suppose $\mu_{\fh}(\ell_1|\phi)$, $\mu_{\fh}(\ell_2|\ell_1)$, $\mu_{\fh}(\ell_2| \phi, \ell_1)$ are high but $ < 1$, which given Proposition \ref{dutch} is the best we can hope for.  By Equation 2  $\mu_{\fh}(\ell_1.\ell_2 |\phi) < \mu_{\fh}(\ell_2 |\phi,\ell)$.  $R$'s probability will thus decrease monotonically as $R$ gets longer, as will the probability of $R$'s last element $\ell_r$. 
Eventually, for some logically valid continuation of $R$.$\ell_{r+1}$, $\mu_{\fh}(\ell_{r+1}|R) < \mu_{\fh}(\psi|R)$ for $\psi$ independent of $R$.  So $\fh$ eventually will surely hallucinate.$\Box$

All LMs that generate responses from a distribution over individual strings
are subject to Proposition \ref{dutch}, and part (ii) applies to logical operators in general: LMs that make reasoning chains will assert invalid ones or will make mistakes in building them, with the probability of logical error increasing as the chain's length increases.  Our theory predicts what many have observed.   


\noindent

Our results show we need to rethink the underlying LM  distribution over strings to solve the  problem of strong hallucinations from negation.  While LMs may have some grasp of negation in simple contexts \cite{gubelmann:handschuh:2022}, this address the underlying problem with an LM's objective function.  We turn to fixing 
that now with a hybrid approach to logical operators.

\section{A hybrid treatment of logical operators for LMs} \label{sec:dynamic}

Logical operators like negation structure meaning recursively by performing a distinctive operation over the content of tokens in its scope.  To develop our hybrid treatment of logical operators $\Lambda$, we adapt this idea and define a logical operator as function transforming distributions over continuations determined by the latent representations of tokens within the scope of those operators into continuations that reflect the semantics of the operator. For example, negation in the formula $\neg A$ transforms distributions over continuations determined
by the latent representation of $A$ into continuations that reflect the semantics of negation and $\neg A$. 

To develop $\Lambda$, we will make use of a dynamic analysis (DS) of logical operators \cite{kamp:1981,kamp:reyle:1993,vaneijk:kamp:1997}.  DS models logical operators as transitions between structures $A$ that are pairs $(U_A,P_A)$.  As \cite{li:etal:2021} have noted, such a pair corresponds to a partition of the tokens in an LM---$U_A$ a set of tokens represents objects and $P_A$ a set of tokens representing properties of those tokens.  We take these tokens to be the output of the processing of some input linguistic context $C$ to the LM.  A second tool we need is the notion of an embedding \cite{chang:keisler:1973}.  An embedding of one structure $A$ derived from linguistic information $C$ into another $B$ from another information source ${\cal V}$ defines satisfaction or truth of $C$ relative to ${\cal V}$.\footnote{Our formulation here echoes maps used in multimodal approaches \cite{li:etal:2019,lu:etal:2019,driess:etal:2023,devillers:etal:2023,vanrullen:kanai:2021}.}  \begin{definition} \label{embed}
$A$ has an {\em embedding} $f: U_A \rightarrow U_B$ in $B$ (written $A \leq_f B) $ iff for each property ascription $\phi \in P_A$ and for all $x_1, ... x_n \in U_A$ if $\phi$ holds of or is {\em satisfied} by  $x_1, ... x_n$ in $A$ then $\phi$ is satisfied by $f(x_1), . . . f(x_n)$ in $B$.  
\end{definition}

 \cite{kamp:reyle:1993} uses embeddings to define the meaning of negation.  Negation structurally takes scope over a representation $A$, and converts it into a property of tokens in a larger structure.  Embeddings and their extensions interpret this structural feature. Suppose for some structure $C$, $A \leq_f C$, we say  $g$ {\em extends} $f$ to an embedding of $B$ in $C$, if: $g \supseteq f$ and $Dom(g) = U_B$ and $B \leq_g C$.  
  \begin{definition} \label{neg} Let $A_\phi$ represent $\phi$. $A_{\neg\phi}$  is satisfied relative to some $\mathfrak{v} \in \mathfrak{V}$ and map $f:$  iff  there is no extension $g$ of $f$ over $U_{A_\phi}$ such that $A_{\phi} \leq_{g} \mathfrak{v}$.
\end{definition}
\noindent

\subsection{Negation as a constraint on continuations}
Pure LMs can't appeal to an external source to characterize negation's effect on their internal representations.  Our key and novel idea uses continuations and their semantics \cite{JPL,asher:etal:2023}.  Continuations $\tau$ of a string $\sigma$ have representations $A_\tau$, where: $U_{A_\sigma} \subseteq U_{A_\tau}$ and $P_{A_\sigma} \subseteq P_{A_\tau}$ (which we write as $A_\sigma \preceq A_\tau$). Then trivially for the identity map $\iota : U_{A_\sigma} \rightarrow U_{A_\sigma}$, $A_\sigma \leq_\iota A_\tau$. Conflating continuations and the structures representing them, a continuation $A_2$ of a context $A_1$ has, given an LM's distribution $\mu$, a probability $\alpha$--written $A_1 \preceq_\alpha A_2$; $\alpha$ is just $\mu(A_2|A_1)$.

Note that $\alpha$ is not the probability of embedding $A_1$ in $A_2$; that probability should always be 1, as $\iota$ always exists and provides the embedding.  While we need to treat embeddings and probabilities when extending our account to quantification, we will not do this here, and simply assume the relevant embedding is $\iota$.

A proper treatment of negation determines how content in the scope of the negation affects coherent or admissible distributions of continuations $A_2$ of a context $A_1$.  
For example, if an object (or token representation) $o$ has the property $\neg A$ in $A_1$, no consistent or coherent continuation of the list of properties for $o$ in $A_2$ should contain $A$.   More precisely, 
Negation imposes three constraints on coherent or {\em admissable} continuations $A_2$ and distributions $\mu$ of an LM relative to a context $A_1$; the fourth constraint below forces $\mu$ to obey the logical interpretation of conjunction or concatenation (.) in strings. Constraint (iv) below ensures that an LM uses conjunction elimination correctly.  For structures $A, B$, let $A + B = (U_A \cup U_B, P_A \cup P_B)$.
\begin{definition} \label{constraint}
Let $A_1 \preceq_{\gamma} A_2$ with $\mu$ an LM.  $A_2$ is an {\em admissible continuation} of $A_1$ relative to $\mu$ only if:\\
(i) If $\{\neg A_k\} \cup P_{A_1}  = P_{A_2}$ with $\mu(A_k | A_1) = \alpha$
 then $\gamma = 1-\alpha$;\\
(ii) if $P_{A_1} = \{\neg A_k\}$ and $A_3 \prec A_2$ and $P_{A_2} = P_{A_1} \cup P_{A_3}$, then\footnote{For details see Appendix A.4.} $\gamma =  \frac{\mu(A_3)(1 - \mu(A_k | A_3))}{1-\mu(A_k)}$;\\ 
 (iii) if $\gamma = 1$, then for any context $C$, $\mu(A_2 |C ) \geq \mu(A_1 |C)$;\\ 
 (iv) for any $A_4 \preceq A_2$ ($A_4$ possibly empty), $\mu(A_1+ A_4|A_2) = 1.$
\end{definition}
\noindent
Definition \ref{constraint} constrains the evolution of an LM's latent representation $A$ and distribution to capture the meanings of logical operators like negation in $A$ in terms of their information update potential, in terms of what distributions over continuations they permit as admissible. 

To illustrate, suppose a distribution $\mu$ with $A_1 \preceq_\gamma A_2$, $A_2$ represents {\em it is not the case that $B$} and $\mu(B|A_1) = \alpha$.   Definition \ref{constraint} says that $A_2$ is an admissible continuation with respect to $\mu$ only if $\gamma = 1 - \alpha$.  Negation maps $\mu(B|A)$ into $1- \mu(B|A)$.  For any context C, unless $\mu(B|C) = .5$, $\mu(B|C) \neq \mu(\neg B |C)$.

Definition \ref{constraint} applies recursively to strings of the form $\neg( \phi . \neg \phi)$ to assign them probability 1 and to $\phi . \neg \phi$ to assign them 0, regardless of the probability of $\phi$.  

We can also model conditionals $A \Rightarrow B$, using Definition \ref{constraint}, by translating {\em if $\phi$ then $\psi$} as $\neg(A_{\phi}.\neg A_{\psi})$.  
So for any context $C$ $\mu(A_{\phi} \Rightarrow A_{\psi}|C) = 1-\mu(A_{\phi}+ A_{\neg\psi}|C)$, which using clause (ii) can be shown to validate modus ponens: i.e., $\mu(B| A_{\phi} \Rightarrow A_{\psi}.A_{\phi}.C) = 1$. So, given a context C like the strong hallucination example $R$ of Section \ref{sec:background}, that contains the conditional {\em (yes to Q1 and no to Q2)} $\Rightarrow$ {\em Entailment} as well as a {\em yes} answer to Q1 and a {\em no} to Q2, then $\mu(\mathit{Entailment}|C) = 1$. Now assuming that our LLM prefers will output the continuation {\em Entailment} over the other possible continuations that have lower probability, we should eliminate such strong hallucinations.

Definition \ref{constraint} also constrains LM objective functions to conform to the truth conditional meanings of not only propositional connectives but quantifiers as well if we complicate Definition \ref{constraint} a bit. Like negation, quantifiers operate on embeddings.  The universal quantifier has the same meaning as that of the conditional we gave above \cite{kamp:reyle:1993,groenendijk:stokhof:1991}, while the meaning of an existential quantifier is captured by the definition of an embedding operating on the elements of $U_A$ that function as variables.  For a few details see Appendix A.5.  

\section{Experiments with our approach on Q\&A, cloze prompts, NLI and negation} \label{sec:empirical}

We illustrate our hybrid approach $\Lambda$ of Section \ref{sec:dynamic} on three applications that feature negation: simple Q\&A, masked knowledge retrieval (MKR) and natural language inference (NLI), each with a different output and input.  We use both encoder and decoder LMs and LLMs.
  
These applications require us to apply $\Lambda$'s constraints logical operators, in particular negation, at various points. Suppose we must predict strings in ${\cal S}$. ${\cal S}$ will have strings or substrings ${\cal S}'$ free of logical operators like negation or conditionals (e.g. simple clauses). $\Lambda$ leverages an LM's distribution $\mu_1$ over ${\cal S}'| {\cal I}'$ where ${\cal I}'$ is also free  operator free and then transforms $\mu_1$ using the constraints in Definition \ref{constraint} into a distribution over ${\cal S|{\cal I}}$, which contain operators.

While $\Lambda$ generalizes to all LLM tasks, our applications only require predicting very short strings. While tasks with long string output are beyond this paper's scope because of the complex evaluations required\cite{li:etal:2023}, we will show how $\Lambda$ can improve the longer reasoning chains of the LLM when prompted on NLI.

\paragraph{Q\&A tasks.}
We examined BERT large and basic encoder models on a question answering task about facts in a synthetic dataset SYN we made.  We used only Y|N questions with positive and negative contexts like:  {\em there was a $<col>$ car} (positive context);
{\em there was no $<col>$ car} (negative context). ``$<col>$'' is a placeholder for a color term.  Given such contexts, we asked, Was there a $<col>$ car? (See Appendix B for details).

Given an input \texttt{\textrm{$<cls>$}\ QUESTION  \textrm{$<sep>$} CONTEXT \textrm{$<sep>$}} and a BERT model, we plotted cosine similarity (cossim) histograms of CLS representations from a positive context and from its corresponding negative context, before and after fine tuning using the training regime of \cite{chaturvedi:etal:2022} on  the CoQA dataset \cite{coqa}.  We used BERT, as its pretraining gives a meaningful CLS representation \cite{bert}.   On pretrained BERT, cossim values ranged between .986 and 1, showing that, as one might expect, pretrained BERT's representations do not give appropriate content to negation.
%

After Q\&A fine tuning, however, cossim values for the CLS representations of positive and negative contexts were much lower, indicating that both BERT models (especially BERT-large with values between 0,34 and 0,38) learned to differentiate positive and negative content in CLS tokens. (Histogram plots are in Appendix C.)

  We then tested fine-tuned BERT and Roberta on our Q\&A task; the large models did perfectly though the small ones mostly gave only a ``no" answer.  The models interpret our task as a binary classification, in which the model considers only two mutually exclusive and exhaustive possible continuations, $P$ (for ``yes'' to $P$?) and $\neg P$ (for ``no'' $P$?. This strategy approximates the truth functional meaning of negation in a simple setting and might account for their success with negative contexts. 
  
  Finally, we used $\Lambda$ and fine-tuned BERT and RoBERTa models only on positive contexts in SYN. 
We accessed the logits, which after softmax give us a probability distribution over the Y|N classes.  Using the probabilities over the positive contexts $(Y|?Q,C)$, we applied Definition \ref{constraint} to get probabilities for outputs with negative responses or contexts--e.g., $(N  | ?Q, C)$ or $(Y | ?Q, \neg C)$. 
$\Lambda$ matched the esults of the large models on our task and yielded the best consistent results for the small models.   

\paragraph{Masked Knowledge Retrieval.}
In a second experiment, we explored a Masked Knowledge Retrieval (MKR), in which the model must compute a sentence completion and lexical item for {\em mask} in context $C$-- where the mask is either in a positive $C$ or negative $\neg C$.  Regardless of the facts, a prediction for $\neg C(\mathit{<mask>})$ should not match the prediction for $C(\mathit{<mask>})$; answers preferred in $C$ should be dispreferred in $\neg C$.

We used  \cite{kassner:schutze:2020}'s (KS) and \cite{jang:etal:2022a}'s (JS) ``negated'' versions of the LAMA dataset \cite{petroni2019language}.  We converted the negative contexts for the MKR task provided in KS and JS into a positive context $C$, which we need for our approach.  KS provided 51 prompt examples, while JS yielded 2926 examples. 

   Pretrained RoBERTa-large and BERT-large had significant numbers of examples with an exact match (EM) for both positive and negative contexts in KS and so showed an inconsistent treatment of negation and strong hallucination.  For instance, given the positive (negative) contexts, {\em A teacher is (not) most likely teaching at a [mask]}, they returned the completions,
  {\em A teacher is most likely teaching at a [school]} and {\em  A teacher is not most likely teaching at a [school]} (See Appendix D for more examples).  Finetuning RoBERTa and BERT on Q\&A as above improved model performance reducing EM but  not eliminating them (see Table 1).  We concluded that fine-tuning on Q\&A does not give a full understanding of negation.
  


 With cloze prompts, an LM originally outputs only the token with the highest score. But an important feature of negation is that it presupposes a set of relevant alternatives \cite{rooth:1992}.   {\em The capital of France is not Marseille} 
 conveys the information that while the capital of France is not Marseille it is some other city or place where people live and work.
 
 To capture the idea of relevant alternatives 
 with $\Lambda$, we modified the LM's output to get its top 5 candidate completions in its distribution.  We then applied Definition \ref{constraint} by computing the probabilities $p_i$ over the top 5 completions  $\sigma_i$ $1 \leq i \leq 5$ of the positive prompt $C(\mathit{<mask>})$; for the negative contexts $\neg C(\mathit{<mask>})$, we assigned $1-p_i$ to $\sigma_i$, reversing the LM's ranking of positive completions.

$\Lambda$ produced $0$ EM on both KS and JS. All completions were meaningful on KS and only relatively few were ungrammatical on JS. Table \ref{cloze-results} gives the performance on MKR of BERT and RoBERTA with pre-training only, fine tuning on Q\&A and $\Lambda$.  

\begin{table}[!h] 
\begin{center}
\small{
\begin{tabular}[2.5in]{|l|llll|}
    \hline
     Model & Dataset & Pre-t& FT-CoQA  & $\Lambda$\\
    \hline
    \multirow{2}{*}{RoBERTa-L}    & KS & 32/51 & 10/51 & 0 / 0 \\
                   & JS & 1038/2926 & 743/2926 & 0/6 \\
                   \hline
\multirow{2}{*}{BERT-L}    & KS & 30/51 & 17/51 & 0/ 0 \\
                   & JS & 970/2926 & 814/2926  & 0/162 \\
 \hline
\end{tabular}
}
\caption{MKR Accuracy for Roberta-large and BERT-large with pre-t(raining  only)/fine-tuned with CoQA (FT-CoQA)  and $\Lambda$. For Pre-tr and FT, we give \#EM / \# examples. For $\Lambda$ we give \#EM /\# non meaningful completions.} \label{cloze-results}
 \end{center}
\end{table}
Using $\Lambda$ on MKR illustrates how our approach forces a re conceptualization of what objective function $\mu(X |C)$ means, where $C$ and $X$ are both finite strings.  $\mu$ is now a measure over the set of all possible continuations of $C$ there are consistent with the substring $C.X$.   For instance, let's look at the example where we have to compute probabilities for the completion of the string {\em the capital of France is not} < mask>, given some possibly empty context $C$.  As we saw that negation is sensitive to alternatives, there should be some restriction to likely alternatives.  But this is what the original distribution over positive data supplies.  Let's suppose that the alternatives in question for the positive context are: (a) Paris, (b) Marseille and (c) Toulouse.  So given that the model is asked what is not the capital of France it must compute probabilities for: ($\neg$ a) {\em the capital of France is not Paris}, ($\neg$ b) {\em the capital of France is not Marseille} and ($\neg$ c) {\em the capital of France is not Toulouse}. Given Definition \ref{constraint}, to compute these probabilities we use what the model  assign the positive strings.  Assume model assigns the following: $\mu (a |C) = .8; \mu (b |C) = .05;
 \mu (c |C) = .15$.   $\mu (\neg a |C) = .2; \mu (\neg b |C) = .95; \mu (\neg c |C) = .85.$  This is what we should expect if the distribution $\mu(X|C)$ measures the set of continuations consistent with $C.X$.  The set of continuations consistent with $C.\neg b$  the set consistent with $C.\neg c$  have a large intersection or overlap.  
 
The space of all possible continuations in a vocabulary $V$  is  $V^\omega \cup V^*$, the set of all finite and countably infinite strings. Each one of the consistent continuations represents a consistent state or world compatible with the context.  The set of continuations defined by $C.\neg b$ is thus a close proxy for the proposition expressed by $C.\neg b$. Given that the training data represents assertions of the agents that produced that data, then each continuation represents an assertible possibility for that collective.  The higher the probability of a given set of continuations, the stronger the degree of assertability.  The higher measure for the $C.\neg b$ set means the continuation $C\neg b$ represents more of the consistent continuations of $C$ than $C.\neg c$.  Our revision of $\mu$ measures this assertability.  $\mu$ now behaves more like an ideal probability function and respects the meanings of logical connectives.  A final point is that higher assertability still guides the model's assertions, so that it guarantees asserting $\neg c$ given $C$ by outputting the string $C.\neg c$.

\paragraph{NLI.}
In a third application, we looked at natural language inference (NLI) with RTE \cite{dagan:etal:2005} and SNLI \cite{bowman:etal:2015}, two datasets modified for negation by \cite{hossain-etal-2020-analysis}.  Each contains a context $C$ and a hypothesis $h$, which are labeled either with  entailment (which we note as $(C,h)$:$E$) or non-entailment $\neg E$ (RTE) or entailment ($E$), contradiction ($Cn$) or neutral ($N$) (SNLI).  \cite{hossain-etal-2020-analysis} negated manually either  $C$ or $h$ or both in portions of RTE and SNLI. Hossain's datasets, $\neg$RTE and $\neg$SNLI, contain no examples of positive cases $(C,h)$.
We reconstructed positive examples (41\% of $\neg$RTE) and for $\neg$SNLI; and to study negation in more detail we completed 117 examples of \cite{hossain-etal-2020-analysis}'s $\neg$SNLI with four inferential patterns: $(C,h)$, $(C,\neg h)$, $(\neg C, h)$ and $(\neg C, \neg h)$. 
 
 In $\neg$SNLI, negations of $(C,h)$ pairs were often wrongly labeled because of non full scope negation.  For instance, there were examples where the original $h$ was of the form {\em a man is smoking} and a context $C$ such that $(C,h):E$ was correct; but where with the negative hypothesis $\neg h$ was {\em a man was not smoking} we had $C,\neg h: Cn$, when $C$ and $\neg h$ were plainly consistent and we should have had $C,\neg h:N$. We made either changed the label or made $h$ and $\neg h$ true contradictories, using a definite description {\em the NP} in $h$ to refer to {\em a NP} from $C$ or replacing {\em a NP} in $h$ with {\em no NP} (for examples see Appendix E).  

In addition our data often featured  less than full sentential scopes for negation on $C$, as well as presuppositional elements (definite noun phrases, adverbial clauses), which scope out of negation.  To deal with this, we divided $C$ into two components $P$ and $C'$, with negation having scope only over $C'$. We then added annotated inferential patterns for both material under the scope of negation and material outside of it. For instance, if we had the 
context {\em a man with blue shoes was not sleeping}
and a hypothesis $h$, we annotated the inference to h from {\em a man with blue shoes} $(P,h)$, and from $h$ to
{\em a man was not sleeping} $(h,C’)$. 
Our datasets containing 2306 annotated inferences are at {\color {magenta} \url{https://github.com/swarnadeep8597/Negation.git}}.

In NLI tasks, the LM must learn to label context and hypothesis with  relations whose definitions involve negation.  This imposes a logical structure on the labels given their intuitive meaning; e.g., entailment (E) between $C$ and $h$ means that there are no situations where $C$ and $\neg h$ hold. From the meaning of entailment and Definition \ref{constraint}, it follows that $\mu(E | C,\neg h) = 1-\mu(E |C,h)$.  Thus, if the model predicts $(C,h)$:$E$, it should predict $(C, \neg h)$: $\neg E$.  To infer labels for $(\neg C, \neg h)$ from positive data, if $(h,C)$:$\neg E/E$, then $(\neg C, \neg h)$: $\neg E/E$. Similarly for $(\neg C, h)$ contexts: if $\neg C, \neg h$:$E$, then $(\neg C, h)$:$\neg E$.   This illustrates the interactions between negation in the data and in the task definition. Our rules are provably correct but incomplete for $\neg$RTE; using only valid rules, we cannot infer $(C,\neg h)$:$E$ from just positive data in the 2 label $E$ and $\neg E$ task. So our method perforce missed those cases for 2 label NLI tasks (see results for $\Lambda$ on $C, \neg h$ in Table \ref{NLI} for $\neg$RTE).  

To account for less than wide scope negations, we developed an algorithm inspired by  \cite{karttunen:peters:1979} that determines values for all configurations based on entailment predictions for $(C,h)$, $(P,h)$ and $(h,C')$.   Algorithms 1 and 2 provide the scoped algorithm for $(\neg C, h)$ and $(\neg C, \neg h)$ in a 2 label NLI problem (the $\neg$RTE dataset) and scoped $C$. As we did not scope $h$, only those cases are relevant.  The algorithm for $\Lambda$ on a 2 label NLI task is in Appendix E.

In our $\neg$SNLI, the three labels $E$, $Cn$ and $N$  also make reference to logical relations involving negation, which  $\Lambda$ must reflect in its algorithm.  
For instance, exploiting these relations for $C,\neg h$ contexts gives: $(C,h)$:$E/Cn/N$ iff $(C,\neg h)$:$Cn/E/N$.  The complete rules for scoped negation
in the 3 label NLI task capture the cases we missed in RTE and are given below.\footnote{We can prove that each of the clauses in the algorithms are correct.  Consider, 
if $(h,C): \neg E$, then $\neg C, \neg h: \neg E$.  
Suppose $(h,C)$: $\neg$ E Then either $h \cap C = 0$ or they overlap but $h \not\subseteq C$ in terms of events or worlds.
In the first case where $h \cap C = 0$, the complement of $C$ contains $h$ and so $(\neg C, \neg h)$: $\neg E$.
In the second case where they overlap but $h \not\subseteq C$, then the complement of $C$ contains some $h$ and some $\neg h$ cases.  So again $(\neg C, \neg h)$: $\neg E$ in this case.}  

\begin{algorithm}
\caption{SNLI Algorithm for $(\neg C,h)$}\label{snli1}
\begin{algorithmic}[1]
\Procedure{Algorithm\-$(\neg C,H)$}{}
\State $(C,h):\gets \text{defined earlier}$
\State $(P,h):\gets \text{defined earlier}$
\State $(h,C'):\gets \text{defined earlier}$
\If {$(C,h):  \textit{E}$}  
    \If {$(P,h):  \textit{E}$}  
        \State $(\neg C,h): \textit{E}$
    \EndIf
    \If {$(P,h):  \textit{N}$ and $(h,C'):  \textit{E}$}  
        \State $(\neg C,h): \textit{$C_n$}$
    \Else
    	\If {$(h,C'):  \textit{N}$}
			\State $(\neg C,h): \textit{N}$
		\EndIf
    \EndIf
\EndIf
\If {$(C,h):  \textit{$C_n$}$}  
    \If {$(P,h):  \textit{$C_n$}$}  
        \State $(\neg C,h): \textit{$C_n$}$
    \EndIf
    \If {$(P,h):  \textit{N}$ and $(h,C'):  \textit{E}$}  
        \State $(\neg C,h): \textit{$C_n$}$
    \Else
    	\If {$(h,C'):  \textit{$C_n$}$}
			\State $(\neg C,h): \textit{E}$
		\EndIf
		\If {$(h,C'):  \textit{N}$}
			\State $(\neg C,h): \textit{N}$
		\EndIf	
    \EndIf
\EndIf
\If {$(C,h):  \textit{N}$ and $(h,C):  \textit{E}$}
	\State $(\neg C,h): \textit{$C_n$}$
\Else
	\State $(\neg C,h): \textit{N}$
\EndIf
\EndProcedure
\end{algorithmic}
\end{algorithm}

\begin{algorithm}
\caption{SNLI Algorithm for $(\neg C,\neg h)$}\label{snli2}
\begin{algorithmic}[1]
\Procedure{Algorithm\-$(\neg C,\neg h)$}{}
\State $(\neg C,h):\gets \text{calculated earlier}$
\If {$(\neg C,h):  \textit{$C_n$}$}  
        \State $(\neg C,\neg h): \textit{$E$}$
\EndIf
\If {$(\neg C,h):  \textit{$E$}$}  
        \State $(\neg C,\neg h): \textit{$C_n$}$
\EndIf
\If {$(\neg C,h):  \textit{$N$}$}  
        \State $(\neg C,\neg h): \textit{$N$}$
\EndIf
\EndProcedure
\end{algorithmic}
\end{algorithm}


 Table \ref{NLI} contains scores for the different configurations of $C$, $h$ and negation.  We first looked at finding a theoretical maximum accuracy for $\Lambda$ using gold labels for for $(C,h)$, $(P,h)$ and $(h, C')$ labels on both our $\neg$RTE and $\neg$SNLI. For our $\neg$RTE, $\Lambda$ yielded an accuracy of 94\% on all $C$,$h$ configurations, while $\Lambda$ basic, an algorithm assuming wide scope negation over $C$, was 11\% less accurate, showing the importance of scoping \cite{kletz:etal:2023}.  
 On our $\neg$SNLI, $\Lambda$ yielded a 96\% accuracy overall.  

Next, we prompted Llama2 7B on our NLI tasks to compute labels for positive $(C,h)$, $(P,h)$ and $(h, C')$ data, to be used with $\Lambda$.  We experimented with different formats; the model did best with prompts that involved using answers to two Y|N questions to infer a label in a chain of thought style \cite{wei:etal:2023}, an example of which is in (R) (Section \ref{sec:background}). Longer prompts for $(C,h)$, shorter ones for $(P,h)$ and $(h, C')$ worked best are given in {\color {magenta} \url{https://github.com/swarnadeep8597/Negation.git}}.

 The prompts in the style of (R) provided extended reasoning chains with conditionals; and Definition \ref{constraint} enabled us to exploit $\Lambda$ also for the positive cases. Llama produced about 12\% incorrect reasoning chains that $\Lambda$ could correct.  However, because the valid inferences led to faulty labels, the scores did not go up.  So we took Llama's labels for the $(C,h)$, $(P,h)$ and $(h, C')$ cases.

\begin{center}
\begin{table}[!h] 
   \small{
\begin{tabular}{|l|l|llll|}
    \hline
    Data & Env. & $C,\neg h$ & $\neg C, h$ & $\neg C, \neg h$ & Full \\
    \hline
    \multirow{4}{*}{$\neg$RTE}   
    & $\Lambda$ basic & .89 & .76 & .91 & .85 \\
    & $\Lambda$ & .89 & .96 & .98 & .94 \\
    & L $\Lambda (.73)$ & .89 & .76 & .76 & .8 \\
    & L (.73) & .76 & .63 & .70 & .71 \\    
    \hline
    \multirow{3}{*}{$\neg$SNLI}   
    & $\Lambda$ & .96 & .97 & .95 & .96 \\
    & L $\Lambda$ (.78) & .77 & .67 & .67 & .72 \\
    & L (.78) & .71 & .12 & .11 & .43 \\
    \hline
\end{tabular}}

\caption{Accuracy on NLI tasks for $\neg$RTE and $\neg$SNLI  datasets.  $\Lambda basic$ accuracies for basic algorithm assuming sentence wide scope. $\Lambda$:  accuracies for the full algorithm with scoping  on the ($\neg$)C,($\neg$)h configurations, given gold labeled $(C,h)$, $(h, C')$, $P,h)$.  L: Llama predictions with best prompts.  L$\Lambda$: predictions using $\Lambda$ given $L$'s predictions for $(C,h)$, $(h, C')$, $P,h)$. \label{NLI}
}
\end{table}
 \end{center}
 
Table \ref{NLI} gives results for (L$\Lambda$) and and Llama alone (L). On our $\neg$RTE, for the positive datasets, L got 73\% correct for $(C,h)$, 76\% correct on $(P,h)$ and 83\% on $(h, C)$ labels.  We used these predictions so that $L, \Lambda$ achieved an average overall accuracy of 80\% on the various $((\neg)C, (\neg) h)$ configurations.   Llama alone (L) had an average accuracy of 71\% on all $((\neg)C, (\neg) h)$ configurations--with 76\% on $(C,\neg h)$, 63\% for $(\neg C, h)$ and 70\%  for $(\neg C, \neg h)$.  $L\Lambda$ significantly outperformed Llama by itself on this task. 
 
 $L$ did less well on our $\neg$SNLI as seen in Table \ref{NLI}. After several attempts we found prompts  on which $L$ predicted 78\% correct on $(C,h)$, 69\% on $(P,h)$ and 70\% on $(h,C')$.  Using these scores $L\Lambda$ gave an overall accuracy of 72\%. $L$ achieved a decent accuracy on $(C, \neg h)$ (accuracy 71\%); but floundered on the $(\neg C, h)$, and $(\neg C, \neg h)$ cases.  $L\Lambda$ far surpassed $L$ on the NLI task with negation. 

We were unsatisfied with the prompting procedures.  Slight variations in prompting produced quite different results, and so we found the procedure quite unstable.  As a result we then fine-tuned Llama2-7B on about half of the original SNLI data set, which only concerns judgments on $C,h$---i.e. positive Context and positive hypothesis. We then gave fined-tuned Llama our $(C,\neg h$, $\neg C, h$ and $\neg C, \neg h$ data sets.  The scores for this model are in the third line of Table \ref{NLI1}.  

We then used $\Lambda$ together with fine tuned Llama's judgments on $C,h$, $h,C$, $h,C'$ and $P,h$ datasets to compute scores for L$\Lambda$.  As can be seen, fine-tuned Llama learned to predict $C,h$ values very well.  It did less well on the data sets with more fragmentary contexts---$P,h$, $h,C$ and $h,C'$.  Still the performance of L$\Lambda$ increased dramatically over that of Llama itself, providing an almost 60\% increase in accuracy and a raw improvement of 27 \% accuracy.

\begin{center}
\begin{table}[!h] 
   \small{
\begin{tabular}{|l|l|llll|}
    \hline
    Data & Env. & $C,\neg h$ & $\neg C, h$ & $\neg C, \neg h$ & Full \\
    {$\neg$SNLI}   
    & $\Lambda$ & .96 & .97 & .95 & .96 \\
    & L $\Lambda$  (.90) & .89 & .65 & .64 & .73 \\
    & L (.90) & .81 & .28 & .27 & .46 \\
    \hline
\end{tabular}}
\caption{Accuracy on NLI tasks f$\neg$SNLI  datasets with Llama fine-tuned on the basic SNLI data set.   L: Llama predictions with fine-tuning on SNLI.  L$\Lambda$: predictions using $\Lambda$ given $L$'s predictions for $(C,h)$, $(h, C')$, $P,h)$. \label{NLI1}
}
\end{table}
 \end{center}


\section{Related work} \label{sec:related}


Definition  \ref{divergence} of strong hallucinations shows they involve necessarily unfaithful content like what \cite{ji:etal:2023} call {\em instrinsic hallucination} in which the unfaithful content contradicts its source.  Strong hallucinations contradict every possible source \cite{moramarco:etal:2022,van-deemter:2024}.  
While researchers have proposed various causes for hallucinations \cite{filippova:2020,parikh:etal:2020,longpre:etal:2021}, we are, as far as we know, the first to define and to analyze strong hallucinations and to derive strong hallucinations from the distribution governing an LM's output.  

Various proposals to avoid hallucinations have also surfaced \cite{nakano:etal:2021,asher:hunter:2022,merrill:etal:2022,gubelmann:handschuh:2022}. The most obvious proposal, building bigger LLMs and larger training corpora, has drawn criticism \cite{filippova:2020,huang:etal:2021,li:etal:2023,goyal:etal:2020,sellam:etal:2020,li:etal:2023}. Our negative results buttress this criticism. Our method for correcting strong hallucinations from negation is formally correct; it's not a form of training but acctually shifts LM output in its last layer. The embeddings we use to define $\Lambda$ could serve to check factual hallucinations as they refine and extend lexical matching and relation extraction techniques \cite{dhingra:etal:2019,cao2018faithful,huang2020knowledge}.


 Prior research \cite{kassner:schutze:2020,hossain-etal-2020-analysis,hosseini:etal:2021,jang:etal:2022a} has reported mediocre results for LM performance with negation on MKR and NLI. 
 Proposition \ref{dutch} provides a formal foundation for these observations, and our tests on pretrained BERT models confirms that they doesn't adequately represent negation. \cite{gubelmann:handschuh:2022} adopt a pragmatic approach that improves LM performance on MKR; our novel, semantic approach gives completely logically coherent MKR performance.  Finally, \cite{truong:etal:2023} investigated GPT style LLMs and found NLMs unable to reason with negation.  Our prompting of Llama2 7B LLM our NLI datasets concurs with their observations.  
 
 \cite{jang:etal:2022} improve LM performance on MKR tasks with negation by training the model on an intermediate task, in which they train an encoder model to predict whether given an input word and sentence the sentence describes it. They then use the model on MKR tasks.  We believe their binary classification task can help understand negation, as our Q\&A experiments indicate.

 Work on neurosymbolic models \cite{poole:2011,deraedt:etal:2020,olausson:etal:2023} is also relevant to our approach. $\Lambda$ leverages the large store of lexical and conceptual knowlwedge in LLMs while constraining the LLM's objective function to comply with the meanings of logical operators.   Other neurosymbolic approaches try to incorporate logical operators into the architecture (using logical gates) without changing the objective function \cite{riegel:etal:2020}.  Our negative results show that such neuro-symbolic approaches are problematic.  More promising post hoc methods use an LLM to translate natural language into some formalism that symbolic methods like theorem provers can manipulate \cite{olausson:etal:2023} or use LLMs to generate additional symbolic rules \cite{kalyanpur:etal:2022}.  However, using LLMs only as translators means we can't access their deep lexical knowledge.  

Our paper's negative results assume a learned distribution over strings from corpora. This is compatible with results about what algorithms an LLM can compute when this distribution is replaced with something else \cite{perez:etal:2021,chiang:etal:2023}. 

\section{Conclusions}

We have investigated strong hallucinations in LMs, originating from faulty meaning representations of logical operators. Strong hallucinations account for logical errors and the failure of generative LLMs as in the example of Section 2 to explain their own predictions even if those predictions are correct.  
We proved 
that under minimal assumptions strong hallucinations must result from an LM's distributions defined over strings.  The received view of LM distributions is thus not tenable. 

Focusing on negation, we proposed a new treatment $\Lambda$ of logical operators, on which 
the operator introduces an operation over latent representations, imposing constraints on how those distributions over those representations evolve in coherent continuations.  Our approach reinterprets an LM's distribution as an assignment of degrees of truth to propositions or sets of strings. The constraints will interact with algorithms for tasks that exploit logical structure.  

We illustrated $\Lambda$ on Q\&A, sentence completion and NLI tasks, with simple and precise evaluation metrics, and showed that it can increase LM performance.  $\Lambda$ only requires LM training on positive data, which is an advantage, since negative data is sparse in real corpora.  For NLI, we tested $\Lambda$ on two new datasets for NLI with negation that modify those in \cite{hossain-etal-2020-analysis}; we showed $\Lambda$ substantially improved LLM performance on NLI label prediction.

 We did not fine tune our models to predict logical operator scope \cite{kletz:etal:2023}.  We also plan to use the newer and more stable Llama 3 models on our NLI task. 
Another task for future work is to deal with the fact that with $\Lambda$ a distribution may generate a high probability for strings that may not be pragmatically relevant or appropriate.  
We will explore how to supplement and refine the distribution with pragmatic constraints that are learned with RLHF methods \cite{mishra:etal:2022,rafailov:etal:2024}. If RLHF has to be done anyway to get good performance from LLMs, perhaps this is not such an onerous step.  We also plan to use new annotated data sets \cite{li:etal:2023} to apply $\Lambda$ to multiple reasoning steps in LMs.  

\section*{Acknowledgments}
{We thank Julie Hunter and Akshay Chaturvedi of the COCOBOT and COCOPIL teams for helpful comments on prior drafts of this paper.  For financial support, we thank the
ANR project COCOBOTS (ANR-21-FAI2-0005), theANR/DGA project DISCUTER (ANR-
21-ASIA-0005), the COCOPIL “Graine” project of the Région Occitanie
of France and the National In-
terdisciplinary Artificial Intelligence Institute AN-
ITI (Artificial and Natural Intelligence Toulouse In-
stitute), funded by the French ‘Investing for the
Future–PIA3’ program under the Grant agreement ANR-19-PI3A-000.  This work was performed using HPC resources from CALMIP (Grant 2016-P23060).

\section{Limitations}

  While we have provided empirical support for our approach $\Lambda$ using both LMs and LLMs on problems with simple and precise evaluation metrics, we haven't deeply investigated negation in full text generation.  We had trouble getting Llama even to produce reasoning chains and explanations of its behavior in NLI. Moreover, while it produced about a 10\% error rate in those reasoning chains, when we corrected the reasoning the conclusions were often faulty because of Llama's faulty inferences on positive NLI contexts.  
  
  $\Lambda$ doesn't get perfect scores on our NLI data because it doesn't capture the complex interactions between scopes of multiple logical operators.   
The scope of operators is crucial as can be seen from our Table \ref{NLI}.  We did not train our models to do scoping.   in future work we will do this following \cite{kletz:etal:2023}.  Preliminary experiments show LLama relatively capable of providing logical forms with plausible scopes for logical operators.

Finally, we need to address the failings of LLMs in dealing with just positive data.  The fine-tuned Llama model performed very well on $(C,h)$ judgments, but on it needs more work to do well on the judgments for $(P,h)$, $(h, C)$ and $(h, C')$.  Our examination of the data on NLI shows that we need a much more comprehensive and varied dataset.  We will continue to enlarge the revised datasets we have provided here in supplementary material.

\bibliography{multimodal}

\begin{thebibliography}{68}
\expandafter\ifx\csname natexlab\endcsname\relax\def\natexlab#1{#1}\fi

\bibitem[{Asher et~al.(2023)Asher, Bhar, Chaturvedi, Hunter, and
  Paul}]{asher:etal:2023}
Nicholas Asher, Swarnadeep Bhar, Akshay Chaturvedi, Julie Hunter, and Soumya
  Paul. 2023.
\newblock Limits for learning with large language models.
\newblock In \emph{12th Joint Conference on Lexical and Computational Semantics
  (*Sem)}. Association for Computational Linguistics.

\bibitem[{Asher and Hunter(2022)}]{asher:hunter:2022}
Nicholas Asher and Julie Hunter. 2022.
\newblock When learning becomes impossible.
\newblock In \emph{2022 ACM Conference on Fairness, Accountability, and
  Transparency}, pages 107--116.

\bibitem[{Asher et~al.(2017)Asher, Paul, and Venant}]{JPL}
Nicholas Asher, Soumya Paul, and Antoine Venant. 2017.
\newblock \href {http://dx.doi.org/10.1007/s10992-016-9402-1} {Message exchange
  games in strategic conversations}.
\newblock \emph{Journal of Philosophical Logic}, 46.4:355--404.

\bibitem[{Baroni and Zamparelli(2010)}]{baroni:zamparelli:2010}
Marco Baroni and Roberto Zamparelli. 2010.
\newblock Nouns are vectors, adjectives are matrices: Representing
  adjective-noun constructions in semantic space.
\newblock In \emph{Proceedings of the 2010 conference on empirical methods in
  natural language processing}, pages 1183--1193.

\bibitem[{Blackburn et~al.(2001)Blackburn, De~Rijke, and
  Venema}]{blackburn2001modal}
Patrick Blackburn, Maarten De~Rijke, and Yde Venema. 2001.
\newblock \emph{Modal logic, Cambridge Tracts in Theoretical Computer Science
  No.53}.
\newblock Cambridge University Press.

\bibitem[{Bowman et~al.(2015)Bowman, Angeli, Potts, and
  Manning}]{bowman:etal:2015}
Samuel~R Bowman, Gabor Angeli, Christopher Potts, and Christopher~D Manning.
  2015.
\newblock A large annotated corpus for learning natural language inference.
\newblock \emph{arXiv preprint arXiv:1508.05326}.

\bibitem[{Cao et~al.(2018)Cao, Wei, Li, and Li}]{cao2018faithful}
Ziqiang Cao, Furu Wei, Wenjie Li, and Sujian Li. 2018.
\newblock Faithful to the original: Fact aware neural abstractive
  summarization.
\newblock In \emph{Proceedings of the AAAI Conference on Artificial
  Intelligence}, volume~32.

\bibitem[{Chang and Keisler(1973)}]{chang:keisler:1973}
Chen~Chung Chang and H~Jerome Keisler. 1973.
\newblock \emph{Model theory}.
\newblock North Holland, Elsevier.

\bibitem[{Chaturvedi et~al.(2024)Chaturvedi, Bhar, Saha, Garain, and
  Asher}]{chaturvedi:etal:2022}
Akshay Chaturvedi, Swarnadeep Bhar, Soumadeep Saha, Utpal Garain, and Nicholas
  Asher. 2024.
\newblock \href {https://doi.org/10.1162/coli_a_00493} {{Analyzing Semantic
  Faithfulness of Language Models via Input Intervention on Question
  Answering}}.
\newblock \emph{Computational Linguistics}, pages 1--37.

\bibitem[{Chiang et~al.(2023)Chiang, Cholak, and Pillay}]{chiang:etal:2023}
David Chiang, Peter Cholak, and Anand Pillay. 2023.
\newblock Tighter bounds on the expressivity of transformer encoders.
\newblock In \emph{International Conference on Machine Learning}, pages
  5544--5562. PMLR.

\bibitem[{Dagan et~al.(2005)Dagan, Glickman, and Magnini}]{dagan:etal:2005}
Ido Dagan, Oren Glickman, and Bernardo Magnini. 2005.
\newblock The pascal recognising textual entailment challenge.
\newblock In \emph{Machine learning challenges workshop}, pages 177--190.
  Springer.

\bibitem[{De~Finetti(1937)}]{definetti:1937}
Bruno De~Finetti. 1937.
\newblock La prévision : ses lois logiques, ses sources subjectives.
\newblock \emph{Annales de l'institut {H}enri {P}oincaré}, 7:1--68.

\bibitem[{De~Raedt et~al.(2020)De~Raedt, Duman{\v{c}}i{\'c}, Manhaeve, and
  Marra}]{deraedt:etal:2020}
Luc De~Raedt, Sebastijan Duman{\v{c}}i{\'c}, Robin Manhaeve, and Giuseppe
  Marra. 2020.
\newblock From statistical relational to neuro-symbolic artificial
  intelligence.
\newblock \emph{arXiv preprint arXiv:2003.08316}.

\bibitem[{Devillers et~al.(2023)Devillers, Mayti{\'e}, and
  VanRullen}]{devillers:etal:2023}
Benjamin Devillers, L{\'e}opold Mayti{\'e}, and Rufin VanRullen. 2023.
\newblock Semi-supervised multimodal representation learning through a global
  workspace.
\newblock \emph{arXiv preprint arXiv:2306.15711}.

\bibitem[{Devlin et~al.(2019)Devlin, Chang, Lee, and Toutanova}]{bert}
Jacob Devlin, Ming-Wei Chang, Kenton Lee, and Kristina Toutanova. 2019.
\newblock \href {http://arxiv.org/abs/1810.04805} {Bert: Pre-training of deep
  bidirectional transformers for language understanding}.

\bibitem[{Dhingra et~al.(2019)Dhingra, Faruqui, Parikh, Chang, Das, and
  Cohen}]{dhingra:etal:2019}
Bhuwan Dhingra, Manaal Faruqui, Ankur Parikh, Ming-Wei Chang, Dipanjan Das, and
  William~W Cohen. 2019.
\newblock Handling divergent reference texts when evaluating table-to-text
  generation.
\newblock \emph{arXiv preprint arXiv:1906.01081}.

\bibitem[{Driess et~al.(2023)Driess, Xia, Sajjadi, Lynch, Chowdhery, Ichter,
  Wahid, Tompson, Vuong, Yu et~al.}]{driess:etal:2023}
Danny Driess, Fei Xia, Mehdi~SM Sajjadi, Corey Lynch, Aakanksha Chowdhery,
  Brian Ichter, Ayzaan Wahid, Jonathan Tompson, Quan Vuong, Tianhe Yu, et~al.
  2023.
\newblock Palm-e: An embodied multimodal language model.
\newblock \emph{arXiv preprint arXiv:2303.03378}.

\bibitem[{Filippova(2020)}]{filippova:2020}
Katja Filippova. 2020.
\newblock Controlled hallucinations: Learning to generate faithfully from noisy
  data.
\newblock \emph{arXiv preprint arXiv:2010.05873}.

\bibitem[{Gaifman(1964)}]{gaifman:1964}
Haim Gaifman. 1964.
\newblock Concerning measures on boolean algebras.
\newblock \emph{Pacific Journal of Mathematics}, (14):61–73.

\bibitem[{Goyal and Durrett(2020)}]{goyal:etal:2020}
Tanya Goyal and Greg Durrett. 2020.
\newblock Evaluating factuality in generation with dependency-level entailment.
\newblock \emph{arXiv preprint arXiv:2010.05478}.

\bibitem[{Groenendijk and Stokhof(1991)}]{groenendijk:stokhof:1991}
Jeroen Groenendijk and Martin Stokhof. 1991.
\newblock \href {https://doi.org/10.1007/BF00628304} {Dynamic predicate logic}.
\newblock \emph{Linguistics and Philosophy}, 14(1):39--100.

\bibitem[{Gubelmann and Handschuh(2022)}]{gubelmann:handschuh:2022}
Reto Gubelmann and Siegfried Handschuh. 2022.
\newblock Context matters: A pragmatic study of plms’ negation understanding.
\newblock In \emph{Proceedings of the 60th Annual Meeting of the Association
  for Computational Linguistics (Volume 1: Long Papers)}, pages 4602--4621.

\bibitem[{Hall(1994)}]{hall:1994}
Ned Hall. 1994.
\newblock Correcting the guide to objective chance.
\newblock \emph{Mind}, 103(412):505--517.

\bibitem[{Hossain et~al.(2020)Hossain, Kovatchev, Dutta, Kao, Wei, and
  Blanco}]{hossain-etal-2020-analysis}
Md~Mosharaf Hossain, Venelin Kovatchev, Pranoy Dutta, Tiffany Kao, Elizabeth
  Wei, and Eduardo Blanco. 2020.
\newblock An analysis of natural language inference benchmarks through the lens
  of negation.
\newblock In \emph{Proceedings of the 2020 Conference on Empirical Methods in
  Natural Language Processing (EMNLP)}, pages 9106--9118.

\bibitem[{Hosseini et~al.(2021)Hosseini, Reddy, Bahdanau, Hjelm, Sordoni, and
  Courville}]{hosseini:etal:2021}
Arian Hosseini, Siva Reddy, Dzmitry Bahdanau, R~Devon Hjelm, Alessandro
  Sordoni, and Aaron Courville. 2021.
\newblock Understanding by understanding not: Modeling negation in language
  models.
\newblock In \emph{Proceedings of the 2021 Conference of the North American
  Chapter of the Association for Computational Linguistics: Human Language
  Technologies}, pages 1301--1312.

\bibitem[{Huang et~al.(2020)Huang, Wu, and Wang}]{huang2020knowledge}
Luyang Huang, Lingfei Wu, and Lu~Wang. 2020.
\newblock Knowledge graph-augmented abstractive summarization with
  semantic-driven cloze reward.
\newblock \emph{arXiv preprint arXiv:2005.01159}.

\bibitem[{Huang et~al.(2021)Huang, Feng, Feng, and Qin}]{huang:etal:2021}
Yichong Huang, Xiachong Feng, Xiaocheng Feng, and Bing Qin. 2021.
\newblock The factual inconsistency problem in abstractive text summarization:
  A survey.
\newblock \emph{arXiv preprint arXiv:2104.14839}.

\bibitem[{Jang et~al.(2022{\natexlab{a}})Jang, Mtumbuka, and
  Lukasiewicz}]{jang:etal:2022a}
Mj~Jang, Frank Mtumbuka, and Thomas Lukasiewicz. 2022{\natexlab{a}}.
\newblock Beyond distributional hypothesis: Let language models learn
  meaning-text correspondence.
\newblock In \emph{Findings of the Association for Computational Linguistics:
  NAACL 2022}, pages 2030--2042.

\bibitem[{Jang et~al.(2022{\natexlab{b}})Jang, Kwon, and
  Lukasiewicz}]{jang:etal:2022}
Myeongjun Jang, Deuk~Sin Kwon, and Thomas Lukasiewicz. 2022{\natexlab{b}}.
\newblock Becel: Benchmark for consistency evaluation of language models.
\newblock In \emph{Proceedings of the 29th International Conference on
  Computational Linguistics}, pages 3680--3696.

\bibitem[{Ji et~al.(2023)Ji, Lee, Frieske, Yu, Su, Xu, Ishii, Bang, Madotto,
  and Fung}]{ji:etal:2023}
Ziwei Ji, Nayeon Lee, Rita Frieske, Tiezheng Yu, Dan Su, Yan Xu, Etsuko Ishii,
  Ye~Jin Bang, Andrea Madotto, and Pascale Fung. 2023.
\newblock Survey of hallucination in natural language generation.
\newblock \emph{ACM Computing Surveys}, 55(12):1--38.

\bibitem[{Kalyanpur et~al.(2022)Kalyanpur, Breloff, and
  Ferrucci}]{kalyanpur:etal:2022}
Aditya Kalyanpur, Tom Breloff, and David~A Ferrucci. 2022.
\newblock Braid: Weaving symbolic and neural knowledge into coherent logical
  explanations.
\newblock In \emph{Proceedings of the AAAI conference on artificial
  intelligence}, volume~36, pages 10867--10874.

\bibitem[{Kamp and Reyle(1993)}]{kamp:reyle:1993}
H.~Kamp and U.~Reyle. 1993.
\newblock \emph{From Discourse to Logic: Introduction to Modeltheoretic
  Semantics of Natural Language, Formal Logic and Discourse Representation
  Theory}.
\newblock Kluwer Academic Publishers.

\bibitem[{Kamp(1981)}]{kamp:1981}
Hans Kamp. 1981.
\newblock A theory of truth and semantic representation.
\newblock In J.\ Groenendijk, T.\ Janssen, and M.\ Stokhof, editors,
  \emph{Formal Methods in the Study of Language}, pages 277--322. Mathematisch
  Centrum, Amsterdam.

\bibitem[{Karttunen and Peters(1979)}]{karttunen:peters:1979}
Lauri Karttunen and Stanley Peters. 1979.
\newblock Conventional lmplicature.
\newblock In \emph{Presupposition}, pages 1--56. Brill.

\bibitem[{Kassner and Sch{\"u}tze(2020)}]{kassner:schutze:2020}
Nora Kassner and Hinrich Sch{\"u}tze. 2020.
\newblock Negated and misprimed probes for pretrained language models: Birds
  can talk, but cannot fly.
\newblock In \emph{Proceedings of the 58th Annual Meeting of the Association
  for Computational Linguistics}, pages 7811--7818.

\bibitem[{Kletz et~al.(2023)Kletz, Candito, and Amsili}]{kletz:etal:2023}
David Kletz, Marie Candito, and Pascal Amsili. 2023.
\newblock Probing structural constraints of negation in pretrained language
  models.
\newblock In \emph{Proceedings of the 24th Nordic Conference on Computational
  Linguistics (NoDaLiDa)}, pages 541--554.

\bibitem[{Lewis(1981)}]{lewis:1981}
David Lewis. 1981.
\newblock A subjectivist’s guide to objective chance.
\newblock In \emph{IFS: Conditionals, Belief, Decision, Chance and Time}, pages
  267--297. Springer.

\bibitem[{Li et~al.(2021)Li, Nye, and Andreas}]{li:etal:2021}
Belinda~Z Li, Maxwell Nye, and Jacob Andreas. 2021.
\newblock Implicit representations of meaning in neural language models.
\newblock In \emph{Proceedings of the 59th Annual Meeting of the Association
  for Computational Linguistics and the 11th International Joint Conference on
  Natural Language Processing (Volume 1: Long Papers)}, pages 1813--1827.

\bibitem[{Li et~al.(2023)Li, Cheng, Zhao, Nie, and Wen}]{li:etal:2023}
Junyi Li, Xiaoxue Cheng, Xin Zhao, Jian-Yun Nie, and Ji-Rong Wen. 2023.
\newblock \href {https://doi.org/10.18653/v1/2023.emnlp-main.397}
  {{H}alu{E}val: A large-scale hallucination evaluation benchmark for large
  language models}.
\newblock In \emph{Proceedings of the 2023 Conference on Empirical Methods in
  Natural Language Processing}, pages 6449--6464, Singapore. Association for
  Computational Linguistics.

\bibitem[{Li et~al.(2019)Li, Yatskar, Yin, Hsieh, and Chang}]{li:etal:2019}
Liunian~Harold Li, Mark Yatskar, Da~Yin, Cho-Jui Hsieh, and Kai-Wei Chang.
  2019.
\newblock Visualbert: A simple and performant baseline for vision and language.
\newblock \emph{arXiv preprint arXiv:1908.03557}.

\bibitem[{Liu et~al.(2019)Liu, Ott, Goyal, Du, Joshi, Chen, Levy, Lewis,
  Zettlemoyer, and Stoyanov}]{roberta:2019}
Yinhan Liu, Myle Ott, Naman Goyal, Jingfei Du, Mandar Joshi, Danqi Chen, Omer
  Levy, Mike Lewis, Luke Zettlemoyer, and Veselin Stoyanov. 2019.
\newblock \href {http://arxiv.org/abs/1907.11692} {Roberta: A robustly
  optimized bert pretraining approach}.

\bibitem[{Longpre et~al.(2021)Longpre, Perisetla, Chen, Ramesh, DuBois, and
  Singh}]{longpre:etal:2021}
Shayne Longpre, Kartik Perisetla, Anthony Chen, Nikhil Ramesh, Chris DuBois,
  and Sameer Singh. 2021.
\newblock Entity-based knowledge conflicts in question answering.
\newblock \emph{arXiv preprint arXiv:2109.05052}.

\bibitem[{Lu et~al.(2019)Lu, Batra, Parikh, and Lee}]{lu:etal:2019}
Jiasen Lu, Dhruv Batra, Devi Parikh, and Stefan Lee. 2019.
\newblock Vilbert: Pretraining task-agnostic visiolinguistic representations
  for vision-and-language tasks.
\newblock \emph{Advances in neural information processing systems}, 32.

\bibitem[{Merrill et~al.(2022)Merrill, Warstadt, and
  Linzen}]{merrill:etal:2022}
William Merrill, Alex Warstadt, and Tal Linzen. 2022.
\newblock Entailment semantics can be extracted from an ideal language model.
\newblock \emph{arXiv preprint arXiv:2209.12407}.

\bibitem[{Mishra et~al.(2022)Mishra, Khashabi, Baral, and
  Hajishirzi}]{mishra:etal:2022}
Swaroop Mishra, Daniel Khashabi, Chitta Baral, and Hannaneh Hajishirzi. 2022.
\newblock Cross-task generalization via natural language crowdsourcing
  instructions.
\newblock pages 3470--3487.

\bibitem[{Moramarco et~al.(2022)Moramarco, Korfiatis, Perera, Juric, Flann,
  Reiter, Savkov, and Belz}]{moramarco:etal:2022}
Francesco Moramarco, Alex~Papadopoulos Korfiatis, Mark Perera, Damir Juric,
  Jack Flann, Ehud Reiter, Aleksandar Savkov, and Anja Belz. 2022.
\newblock Human evaluation and correlation with automatic metrics in
  consultation note generation.
\newblock In \emph{ACL 2022: 60th Annual Meeting of the Association for
  Computational Linguistics}, pages 5739--5754. Association for Computational
  Linguistics.

\bibitem[{Nakano et~al.(2021)Nakano, Hilton, Balaji, Wu, Ouyang, Kim, Hesse,
  Jain, Kosaraju, Saunders et~al.}]{nakano:etal:2021}
Reiichiro Nakano, Jacob Hilton, Suchir Balaji, Jeff Wu, Long Ouyang, Christina
  Kim, Christopher Hesse, Shantanu Jain, Vineet Kosaraju, William Saunders,
  et~al. 2021.
\newblock Webgpt: Browser-assisted question-answering with human feedback.
\newblock \emph{arXiv preprint arXiv:2112.09332}.

\bibitem[{Olausson et~al.(2023)Olausson, Gu, Lipkin, Zhang, Solar-Lezama,
  Tenenbaum, and Levy}]{olausson:etal:2023}
Theo~X Olausson, Alex Gu, Ben Lipkin, Cedegao~E Zhang, Armando Solar-Lezama,
  Joshua~B Tenenbaum, and Roger~P Levy. 2023.
\newblock Linc: A neurosymbolic approach for logical reasoning by combining
  language models with first-order logic provers.
\newblock In \emph{The 2023 Conference on Empirical Methods in Natural Language
  Processing}.

\bibitem[{Parikh et~al.(2020)Parikh, Wang, Gehrmann, Faruqui, Dhingra, Yang,
  and Das}]{parikh:etal:2020}
Ankur~P Parikh, Xuezhi Wang, Sebastian Gehrmann, Manaal Faruqui, Bhuwan
  Dhingra, Diyi Yang, and Dipanjan Das. 2020.
\newblock Totto: A controlled table-to-text generation dataset.
\newblock \emph{arXiv preprint arXiv:2004.14373}.

\bibitem[{P{\'e}rez et~al.(2021)P{\'e}rez, Barcel{\'o}, and
  Marinkovic}]{perez:etal:2021}
Jorge P{\'e}rez, Pablo Barcel{\'o}, and Javier Marinkovic. 2021.
\newblock Attention is turing-complete.
\newblock \emph{Journal of Machine Learning Research}, 22(75):1--35.

\bibitem[{Petroni et~al.(2019)Petroni, Rockt{\"{a}}schel, Miller, Lewis,
  Bakhtin, Wu, and Riedel}]{petroni2019language}
F.~Petroni, T.~Rockt{\"{a}}schel, A.~H. Miller, P.~Lewis, A.~Bakhtin, Y.~Wu,
  and S.~Riedel. 2019.
\newblock Language models as knowledge bases?
\newblock In \emph{In: Proceedings of the 2019 Conference on Empirical Methods
  in Natural Language Processing (EMNLP), 2019}.

\bibitem[{Poole(2011)}]{poole:2011}
David Poole. 2011.
\newblock Logic, probability and computation: Foundations and issues of
  statistical relational ai.
\newblock In \emph{International Conference on Logic Programming and
  Nonmonotonic Reasoning}, pages 1--9. Springer.

\bibitem[{Rafailov et~al.(2024)Rafailov, Sharma, Mitchell, Manning, Ermon, and
  Finn}]{rafailov:etal:2024}
Rafael Rafailov, Archit Sharma, Eric Mitchell, Christopher~D Manning, Stefano
  Ermon, and Chelsea Finn. 2024.
\newblock Direct preference optimization: Your language model is secretly a
  reward model.
\newblock \emph{Advances in Neural Information Processing Systems}, 36.

\bibitem[{Ramsey(1931)}]{ramsey:1931}
Frank~Plumpton Ramsey. 1931.
\newblock \emph{The foundations of mathematics and other logical essays}.
\newblock K. Paul, Trench, Trubner \& Company, Limited.

\bibitem[{Reddy et~al.(2019)Reddy, Chen, and Manning}]{coqa}
Siva Reddy, Danqi Chen, and Christopher~D. Manning. 2019.
\newblock \href {https://doi.org/10.1162/tacl_a_00266} {{CoQA: A Conversational
  Question Answering Challenge}}.
\newblock \emph{Transactions of the Association for Computational Linguistics},
  7:249--266.

\bibitem[{Riegel et~al.(2020)Riegel, Gray, Luus, Khan, Makondo, Akhalwaya,
  Qian, Fagin, Barahona, Sharma et~al.}]{riegel:etal:2020}
Ryan Riegel, Alexander Gray, Francois Luus, Naweed Khan, Ndivhuwo Makondo,
  Ismail~Yunus Akhalwaya, Haifeng Qian, Ronald Fagin, Francisco Barahona, Udit
  Sharma, et~al. 2020.
\newblock Logical neural networks.
\newblock \emph{arXiv preprint arXiv:2006.13155}.

\bibitem[{Rooth(1992)}]{rooth:1992}
M.\ Rooth. 1992.
\newblock A theory of focus interpretation.
\newblock \emph{Natural Language Semantics}, 1(1):75--116.

\bibitem[{Sellam et~al.(2020)Sellam, Das, and Parikh}]{sellam:etal:2020}
Thibault Sellam, Dipanjan Das, and Ankur~P Parikh. 2020.
\newblock Bleurt: Learning robust metrics for text generation.
\newblock \emph{arXiv preprint arXiv:2004.04696}.

\bibitem[{Talmor et~al.(2020)Talmor, Elazar, Goldberg, and
  Berant}]{talmor:etal:2020}
Alon Talmor, Yanai Elazar, Yoav Goldberg, and Jonathan Berant. 2020.
\newblock olmpics-on what language model pre-training captures.
\newblock \emph{Transactions of the Association for Computational Linguistics},
  8:743–758.

\bibitem[{Tarski(1944)}]{tarski:1944}
Alfred Tarski. 1944.
\newblock The semantic conception of truth: and the foundations of semantics.
\newblock \emph{Philosophy and phenomenological research}, 4(3):341--376.

\bibitem[{Tenney et~al.(2019)Tenney, Xia, Chen, Wang, Poliak, McCoy, Kim,
  Van~Durme, Bowman, Das et~al.}]{tenney:etal:2018}
Ian Tenney, Patrick Xia, Berlin Chen, Alex Wang, Adam Poliak, R~Thomas McCoy,
  Najoung Kim, Benjamin Van~Durme, Samuel~R Bowman, Dipanjan Das, et~al. 2019.
\newblock What do you learn from context? probing for sentence structure in
  contextualized word representations.
\newblock In \emph{International Conference on Learning Representations}.

\bibitem[{Truong et~al.(2023)Truong, Baldwin, Verspoor, and
  Cohn}]{truong:etal:2023}
Thinh~Hung Truong, Timothy Baldwin, Karin Verspoor, and Trevor Cohn. 2023.
\newblock Language models are not naysayers: An analysis of language models on
  negation benchmarks.
\newblock \emph{arXiv preprint arXiv:2306.08189}.

\bibitem[{van Deemter(2024)}]{van-deemter:2024}
Kees van Deemter. 2024.
\newblock The pitfalls of defining hallucination.
\newblock \emph{Computational Linguistics}, pages 1--10.

\bibitem[{van Eijck and Kamp(1997)}]{vaneijk:kamp:1997}
J.\ van Eijck and H.\ Kamp. 1997.
\newblock Representing discourse in context.
\newblock In Johan {van Benthem} and Alice {ter Meulen}, editors,
  \emph{Handbook of Logic and Linguistics}, pages 179--237. Elsevier.

\bibitem[{VanRullen and Kanai(2021)}]{vanrullen:kanai:2021}
Rufin VanRullen and Ryota Kanai. 2021.
\newblock Deep learning and the global workspace theory.
\newblock \emph{Trends in Neurosciences}, 44(9):692--704.

\bibitem[{Vaswani et~al.(2017)Vaswani, Shazeer, Parmar, Uszkoreit, Jones,
  Gomez, Kaiser, and Polosukhin}]{vaswani2017attention}
Ashish Vaswani, Noam Shazeer, Niki Parmar, Jakob Uszkoreit, Llion Jones,
  Aidan~N Gomez, {\L}ukasz Kaiser, and Illia Polosukhin. 2017.
\newblock Attention is all you need.
\newblock \emph{Advances in neural information processing systems}, 30.

\bibitem[{Wei et~al.(2023)Wei, Wang, Schuurmans, Bosma, Ichter, Xia, Chi, Le,
  and Zhou}]{wei:etal:2023}
Jason Wei, Xuezhi Wang, Dale Schuurmans, Maarten Bosma, Brian Ichter, Fei Xia,
  Ed~Chi, Quoc Le, and Denny Zhou. 2023.
\newblock \href {http://arxiv.org/abs/2201.11903} {Chain-of-thought prompting
  elicits reasoning in large language models}.

\bibitem[{Zhu and Rudzicz(2020)}]{zhu:rudizicz:2020}
Zining Zhu and Frank Rudzicz. 2020.
\newblock An information theoretic view on selecting linguistic probes.
\newblock \emph{arXiv preprint arXiv:2009.07364}.

\end{thebibliography}

\appendix

\section*{Appendix}
{\bf A.1 Language, probabilities and propositions \ref{basic-prob} and \ref{log-equiv}}

To show the connection between probability and logical truth, we will use the notion of a modal model \cite{blackburn2001modal}.  Modal models consist of a set of points of evaluation and an interpretation function assigning appropriate meanings to tokens. Let $\mathfrak{A}$ be a modal model \cite{blackburn2001modal} with a set of possible worlds or points of evaluation $W_{\mathfrak{A}}$. For $w \in W_{\mathfrak{A}}$, we write $\mathfrak{A}, w \models \phi$ for the fact that $\phi$ is true in $\mathfrak{A}$ at $w$, and $\|\phi\|_{\mathfrak{A}} = \{w \in W_{\mathfrak{A}}: \mathfrak{A}, w \models \phi\}$.  
\begin{definition} \label{lc}
Let $\phi, \psi \in Sent({\cal L})$.  $\phi$ is a semantic or logical consequence of $\psi$ ($\psi \models \phi$) if for all models $\mathfrak{A}$ and worlds $w \in W_{\mathfrak{A}}$ such that $\mathfrak{A},w \models \psi$, then
$\mathfrak{A},w \models \phi$ \cite{chang:keisler:1973}. 
\end{definition}
 \begin{prop}\label{popper} Let $P$ be a probability distribution that respects the $\sigma$ algebra $\mathbb{P}$.  Then $P$ determines a probability distribution $\pi$ over $Sent({\cal L})$ such that the following axioms A hold: $\pi(\neg \phi) = 1 -\pi(\phi)$; If $\models \phi$, then $\pi(\phi) = 1$; 
if $\phi \models \psi$, then $\pi(\phi) \leq \pi(\psi)$; if $\models \neg(\phi \wedge \psi)$, then  $\pi(\phi \vee \psi) = \pi(\phi) + \pi(\psi)$; $\pi(\phi \wedge \psi) = \pi(\|\phi \| \cap \| \psi \|)$, and $\pi (\forall x \phi) = lim_{n \rightarrow \infty} \bigwedge_{a_1,...a_n \in C_{\cal L}} \pi(\phi(\frac{x}{a}))$.  
\end{prop}

A straightforward consequence Proposition \ref{popper} is that  $\pi$  must assign the same probability to sentences $\phi, \psi$ if $\models \phi \leftrightarrow \psi$ (are logically equivalent). 

Finally, we can relate probability distributions $\pi$ to models of linguistic meaning.  Let ${\cal L}$ be a first order language.
\begin{prop} \label{basic-prob} 
(i) Every probability function $\pi: Sent({\cal L}) \rightarrow [0, 1]$ that respects $\mathbb{P}$ defines a modal model $\mathfrak{A}$ of ${\cal L}$ with worlds $W$ where: if $\phi$ is a logical truth, $\|\phi\|_{\mathfrak{A}} = W$; if $\phi$ is inconsistent, then $\|\phi\|_{\mathfrak{A}} = \emptyset$; If $\models \phi \leftrightarrow \psi$, then $\| \phi \|_{\mathfrak{A}} = \|\psi \|_{\mathfrak{A}}$ and all identities of $\mathbb{P}$ hold in ${\cal A}$; finally if $\phi \models \psi$, $\pi(\psi | \phi) = 1$. (ii) Every model of ${\cal L}$ defines a probability function $\pi: Sent({\cal L}) \rightarrow [0, 1]$ verifying the axioms $A$ of Proposition \ref{popper}, and (iii) if $\pi : Sent({\cal L}) \rightarrow [0, 1]$ does not verify the axioms $A$, it does not define a model of ${\cal L}$, or preserve semantic consequence.
 
\end{prop}

Proof: To prove (i), we consider a set of worlds $W$ where $\pi$ provides a uniform distribution over $w \in W$.  
If $\pi(\phi) = \alpha$, we build a model $\mathfrak{A}$ such that where $\| \phi \| = \{ w \in W : \mathfrak{A}, w \models \phi \}$,  $\sum_{w \in \| \phi \|} \pi(w) = \alpha$.   Now suppose $\phi$ is a logical truth.  Then $\pi(\phi) = 1$.  So $\sum_{w \in \| \phi \|} \pi(w) = 1$.  Given that $\pi$ is a probability measure over $W$, $\| \phi \| = W$.  Similarly if $\phi$ is inconsistent, then $\pi(\phi) = 0$.  And so $\| \phi \| = \emptyset$.  Given the axioms in A,  if $\models \phi \rightarrow \psi$, then $\pi(\phi) \leq \pi(\psi)$; we then set $\| \phi\| \subseteq \| \psi \|$ in $\mathfrak{A}$.  So if $\models \phi \leftrightarrow \psi$, $\| \phi\| = \| \psi \|$.  This means that all the identities of $\mathbb{P}$ and all logical equivalences hold in $\mathfrak{A}$.  Finally, if $\phi \models \psi$, $\|\phi \wedge \psi\| = \|\psi\|$ given our definition of $\mathfrak{A}$. Then $\pi(\phi |\psi) = \frac{\pi(||\phi || \ \cap ||\psi||)}{\pi (||\psi||)} = \frac{\pi(||\psi||)}{\pi (||\psi||)} = 1$.\\ To prove (ii),
assume a modal model $\mathfrak{A}$ and define a set of measures $\Pi$ on $W_{\mathfrak{A}}$ such that $\pi \in \Pi$, $\pi(W_{\mathfrak{A}}) = 1$; $\pi(\emptyset) = 0$; and for $p, q \subseteq W_{\mathfrak{A}}$, if $p \subset q$ then $\pi(p) \leq \pi(q)$; and if $p \cup q =0$, then $\pi(p \cup q ) = \pi(p) \cup \pi(q)$.  Such measures exist \cite{gaifman:1964} and in virtue of Proposition \ref{boolean} obey axioms $A$.\\
To prove (iii), if $\pi$ does not obey axioms $A$, then one of the clauses of satisfaction for {\cal L} formulas \cite{tarski:1944} will fail; since a model of ${\cal L} $ must by definition satisfy those clauses, $\pi$ cannnot define a model.
$\Box$

\begin{prop} \label{log-equiv}
Every ideal distribution defines a modal model of ${\cal L}$.
\end{prop}
\noindent
Proof: By hypothesis the objective probability distribution $P$ respects the structure of $\mathbb{P}$ and thus $P$ obeys the axioms in $A$.  By Proposition \ref{basic-prob} $P$ defines a model ${\mathfrak A}_P$ of ${\cal L}$. By definition of $\kappa$, $\kappa(\phi | T) = P(\phi)$, so $\kappa( . | T)$ defines a model ${\mathfrak A}_\kappa$ of ${\cal L}$ that is elementarily equivalent to ${\mathfrak A}_P$ \cite{chang:keisler:1973}. But this means that $\kappa$ must respect the structure of $\mathbb{P}$ and has the requisite properties via Proposition \ref{basic-prob} to define a model 

\noindent
{\bf A.2 Dutch books \label{dutchbooks}} \\

There is an intuitive link between betting behavior and degrees of belief.  For instance, if I believe $p$ to degree $.5$, if I am rational I will bet that $p$ is true only given even odds; but if my degree of belief is $.6$, then I will take the bet given 4/6 odds or greater (I will receive 4 euros on winning and will pay out 6 euros if I lose). Linking probability to betting behavior leads to the idea of a {\em Dutch book}. A Dutch book is a set of odds and bets, established by a bookmaker, that ensures that the bookmaker will profit at the expense of the gambler no matter what the facts are.  

We predict that LMs should be subject to Dutch Books.  Given Proposition \ref{dutch}, we have seen that under minimal conditions,  $\mu(\neg \phi) + \mu( \phi) < 1$ and $\mu( \phi \vee \neg \phi) < 1$ or that $\mu(\phi) \neq \mu(\psi)$, although $\phi$ and $\psi$ are logically equivalent.  So a bookie can offer $\fh$ bets that on the set of possibilities  $D = \{\phi, \neg \phi\}$.  Given $\fh$'s probabilities, $\fh$ should bet that none of $D$'s possibilities holds only if the bookie gives $\fh$ odds reflecting $1 - \mu(\neg \phi) + \mu( \phi)$.  Since this is positive, $\fh$ will take a bet the bookie can't lose and so $\fh$ will necessarily lose money. A bookie can also take advantage of $\fh$, given that in Proposition \ref{dutch} that there are semantically equivalent $\phi$ and $\psi$ such that $\mu(\phi) \neq \mu(\psi)$.  We prompted ChatGPT to bet on semantically equivalent, short sentences (for an interaction see Appendix G).  ChatGPT failed to recognize semantic equivalences and so was subject to Dutch books. 

Note that $\fh$ can have a distribution where it is not induced to affirm both $\phi$ and $\neg \phi$ but still be subject to a Dutch book argument.  Consider a  language $L$ with just four sentences or strings $L = \{p, \neg p, q, p \vee \neg p\}$, and let $\mu_{\fh}(p) =\mu_{\fh}(q) = \frac{1}{4}$ with $\mu_{\fh}(\neg p) = 0$.  This will not induce $\fh$ by our definitions to affirm an outright contradiction.  The incoherence is more subtle.  What is the probability that $\fh$ could assign to $p \vee \neg p$ in this case?  Since $\sum_{\phi \in L} \mu(\phi) = 1$, 
$\mu(p \vee \neg p) = .5$ as a maximum.  So the bookie can offer $\fh$ an astronomical reward to bet against $p \vee \neg p$, and $\fh$ should accept the bet if it's maximizing its expected gain.  /But inevitably, $\fh$ will lose money.

i) 

{\bf A.4 Explanation of constraint in admissable continuations}

\begin{equation} P(a | b^c) = \frac{P(a \cap b^c)}{P(b^c)}\end{equation}
where $b^c$ is the complement of $b$.
\begin{equation} \label{basic} (a \cap b^c) \cup (a \cap b) = a
\end{equation}
Since the 2 sets on the left hand side of (\ref{basic}) are disjoint, 
$P(a) = P(a \cap b^c) + P( a \cap b)$, so $P( a \cap b^c)  = P(a) - P( a \cap b)$.
Putting that together with what we had above:\\
\begin{equation}\frac{P(a \cap b^c)}{P(b^c)}= \frac{P(a) - P(a \cap b)}{1 - P(b)} \end{equation}

 \begin{equation}P(a | b^c) = \frac{P(a) - P(b|a).P(a)}{1 - P(b)}\end{equation}
 \begin{equation}P(a | b^c) =  \frac{P(a)(1 - P(b|a))}{1-P(b)}\end{equation}
 
 {\bf A.5 Quantifiers in our continuation semantics}  Like negation, quantifiers operate on embeddings.  The universal quantifier has the same meaning as that of the conditional we gave above \cite{kamp:reyle:1993,groenendijk:stokhof:1991}, while the meaning of an existential quantifier is captured by the definition of an embedding operating on the elements of $U_A$ that function as variables.  For example, suppose $A_1, A_2$ as in Definition \ref{constraint} and suppose {\em Every}$(A_j, A_k) \in P_{A_2}$. In our continuation semantics this formula has the following meaning:  for every embedding $f$  $A_2 \leq_{f,\alpha} A_j$ there is an embedding $g \supset f$ and $\beta \geq \alpha$ such that  $A_k \leq_{f, \beta} A_2$.  This is equivalent to the meaning of $A_i \Rightarrow A_k$\cite{kamp:reyle:1993}.

{\bf B: Schemas Used to Construct Our synthetic dataset SYN}\\
We design two datasets, one containing positive and another containing the negative samples.\\
The positive schemas are as follows :-
\begin{enumerate}
    \item CONTEXT: There was a $<col>$ car. QUESTION: Was there a $<col>$ car ?
    \item CONTEXT: John played with a $<col>$ ball. \\QUESTION: Did john play with a $<col>$ ball ?
    \item CONTEXT: The man was wearing a $<col>$ shirt. \\ QUESTION: Did the man wear a $<col>$ shirt ?
    \item CONTEXT: The house had a $<col>$ window. \\ QUESTION: Did the house have a $<col>$ window ?
    \item CONTEXT: A $<col>$ glass was on the table. \\ QUESTION: Was there a $<col>$ glass on the table?
\end{enumerate}
The negative schemas are given by :-
\begin{enumerate}
    \item CONTEXT: There was no $<col>$ car. QUESTION: Was there a $<col>$ car ?
    \item CONTEXT: John played with no $<col>$ ball. \\QUESTION: Did john play with a $<col>$ ball ?
    \item CONTEXT: The man was wearing no $<col>$ shirt. \\ QUESTION: Did the man wear a $<col>$ shirt ?
    \item CONTEXT: The house had no $<col>$ window. \\ QUESTION: Did the house have a $<col>$ window ?
    \item CONTEXT: No $<col>$ glass was on the table. \\ QUESTION: Was there a $<col>$ glass on the table?
\end{enumerate}

We generate different data points by choosing $<col>$ from a list of colors (red,blue,green,yellow,black,white)\\

{\bf C.Histograms Showing that Specialised training on QA teaches the models something about negation:}

\begin{figure}[h]
\centering
\includegraphics[width=0.8\linewidth]{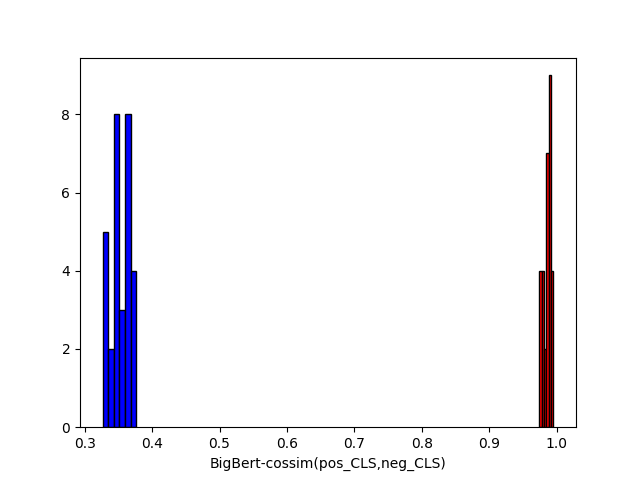}
\includegraphics[width=0.8\linewidth]{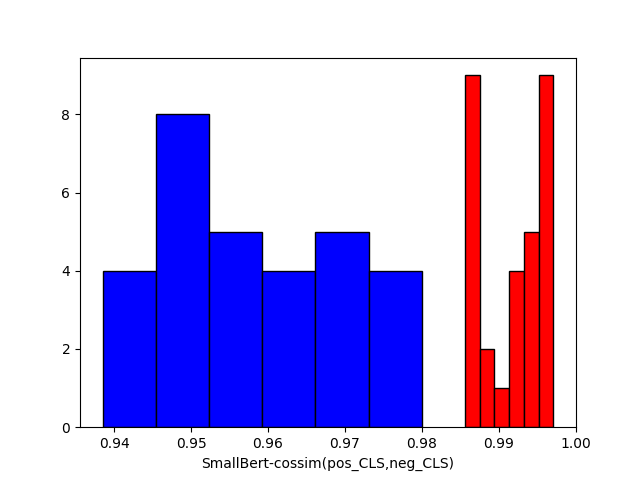}
\caption{Histogram plot of cossim values for positive and negative contexts with BERT-large (above) BERT-base (below) before (red) and after fine-tuning (blue).}
\label{fig:gpt}
\end{figure}
The values in red, the cossim values of CLS tokens without fine-tuning are on the higher end of the historgram, indicating a high cosssim between positive and negative samples.

\hidden{{\bf D. Model behavior on SYN and SIM datasets.}

The smaller LM models, BERT and RoBERTa basic, were worse on the positive contexts than the negative ones, tending to answer ``no'' more often than not. They also sometimes got confused by non-typical color choices (given the context, {\em John played with a black ball}, the base models sometimes answered {\em no} to {\em did John play with a black ball?}.  We separated out a subset SIM of SYN, where we filled the schemas in with typical colors and other properties of the objects in question.  The smaller models far fewer difficulties with the (Sim) examples; this suggests that their mediocre performance on SYNis not about negation but rather about its pretraining inducing it to reject non standard properties, like non standard colors for bats and balls and the like.}

{\bf D.Some Examples from Cloze Prompts Showing Wrong Predictions from LMs without our negation constraint and the change due to our negation constraint.}\\
We show the positive and negative strings predicted by out of the box,\\ \textbf{bert-base}
\begin{enumerate}
    \item A teacher is most likely teaching at a [school].\\
          A teacher is not most likely teaching at a [school].
    \item The Teatr Wielki is a [museum]\\
          The Teatr Wielki is not a [museum]
    \item The LDS Church focuses on [individual] mentorship.\\
          The LDS Church does not focus on [individual] mentorship.
    \item Warsaw is the most diverse [city] in Poland.\\
          Warsaw is not the most diverse [city] in Poland.
    \item The 1893 World's Columbian Exposition was held in [chicago].\\
          The 1893 World's Columbian Exposition was not held in [chicago].
\end{enumerate}

With our approach, we look at the top 5 probabilities which are predicted for the positive string and reverse them (1-P), this gives us the following [MASK] completions,
\begin{enumerate}
    \item A teacher is not most likely teaching at a [hospital].
    \item The Teatr Wielki is not a [club]
    \item The LDS Church does not focus on [community] mentorship.
    \item Warsaw is not the most diverse [suburb] in Poland.
    \item The 1893 World's Columbian Exposition was not held in [philadelphia].  
\end{enumerate}
\textbf{roberta-large}
\begin{enumerate}
    \item Quran is a <religious> text.\\
          Quran is not a <religious> text.
    \item Isaac's chains made out of <wood>.\\
          Isaac's chains made out of <wood>.
    \item The sporting capital of Australia is <Melbourne>.\\
          The sporting capital of Australia is not <Melbourne>.
    \item Warsaw is the most diverse [city] in Poland.\\
          Warsaw is not the most diverse [city] in Poland.
    \item The 1893 World's Columbian Exposition was held in [chicago].\\
          The 1893 World's Columbian Exposition was not held in [chicago].
\end{enumerate}
Our approach on the other hand gives:\\
\begin{enumerate}
    \item Quran is not a <biblical> text.
    \item Isaac's chains made out of <glass>.
    \item The sporting capital of Australia is not <Brisbane>.
    \item Warsaw is not the most diverse [place] in Poland.
    \item The 1893 World's Columbian Exposition was not held in [philadelphia].  
\end{enumerate}

We assume that the model can generate correct completions for the positive sentences,which when reversed might give us a likely completion for the negative string. \\

{\bf E. SNLI examples and corrections}

Here is a typical example from SNLI:\\
(C) "Tattooed young woman chains her bicycle to a signpost while juggling a guitar on her back",\\
($\neg h$) "A woman with tattoos does not chain her bike to a post",\\
 
 which is (incorrectly) labelled with contradiction, when the judgment should be neutral due to the presence of the two indefinites.  To get make $\neg h$ a contradictory of $C$, we had two strategies:

 "Tattooed young woman chains her bicycle to a signpost while juggling a guitar on her back",\\
 "No woman with tattoos chains her bike to a post",\\

 and 
 "Tattooed young woman chains her bicycle to a signpost while juggling a guitar on her back",\\
 "The woman with tattoos did not chain her bike to a post",\\

 Both of these devices restore the desired contradictory status and contrast nicely with the original entailment pair from SNLI:

"Tattooed young woman chains her bicycle to a signpost while juggling a guitar on her back",\\
 "A woman with tattoos chains her bike to a post",\\

{\bf F: Algorithms for NLI} 

{\bf Algorithms for $\neg$RTE:}
We here give the full unscoped rules for NLI on $\neg$RTE.

\noindent
For $(C \neg h)$.\\
(i) If $(C,h)$:$E$, then $(C, \neg h)$: $\neg E$.  \\
(ii) If $(C,h)$:$\neg E$ and if $(h,C)$:$\neg E$, then $(C, \neg h)$: $\neg E$.\\

\noindent
For $(\neg C, \neg h)$.\\
If $(h,C)$:$\neg E/E$, then $(\neg C, \neg h)$: $\neg E/E$.

\noindent
For $(\neg C, h)$.\\ 
(i) if $\neg C, \neg h$:$E$, then $(\neg C, h)$:$\neg E$. \\
(ii) If $[(C,h)$:$\neg E$ and $(h, C)$:$E$, then $(C, \neg h)$:$\neg E$.  \\
(iii) If $\{[(C,h)$:$\neg E$ and $(h, C)$:$ E]$ or $[(C,h)$:$ E$ and $(h, C)$:$\neg E]\}$, then $(\neg C, h)$:$\neg E$. Else, $(\neg C, (\neg)h)$: $\neg E$ .

Here  are the full scoped rules for the two label ($E , \neg E)$ problem.

\makeatletter
\def\BState{\State\hskip-\ALG@thistlm}
\makeatother
\begin{algorithm} \label{RTE11}
\caption{RTE Algorithm for $(\neg C,h)$} 
\begin{algorithmic}[1]
\Procedure{Algorithm\-$(\neg C,H)$}{}
\State $(C,h):\gets \text{defined in paper}$
\State $(P,h):\gets \text{P defined in paper}$
\If {$(h,C):  \textit{$\neg E$}$}  
    \If {$(C,h):  \textit{E}$ and $(P,h):  \textit{E}$}  
        \State $(\neg C,h): \textit{E}$
    \Else  
        \State $(\neg C,h): \textit{$\neg E$}$
    \EndIf
\EndIf
\If {$(h,C):  \textit{E}$} {$(\neg C,h)$: \textit{$\neg E$}}
\EndIf
\EndProcedure
\end{algorithmic}
\end{algorithm}

\begin{algorithm}
\caption{RTE Algorithm for $(\neg C,\neg h)$}\label{RTE21}
\begin{algorithmic}[1]
\Procedure{Algorithm\-$(\neg C,\neg H)$}{}
\State $(C,h):\gets \text{C,h defined in paper}$
\State $(P,h):\gets \text{P defined in paper}$
\State $(h,C'):\gets \text{C' defined in paper}$
\If {$(h,C):  \textit{E}$}  
    \If {$(P,h):  \textit{E}$}  
        \State $(\neg C,h): \textit{E}$
    \Else  
        \State $(\neg C,h): \textit{$\neg E$}$
    \EndIf
\EndIf
\If {$(h,C):  \textit{$\neg E$}$}
	\If {$(h,C'):  \textit{E}$}
		\State $(\neg C,h): \textit{E}$
	\EndIf
\EndIf
\EndProcedure
\end{algorithmic}
\end{algorithm}




}

XXXLet $\neg C = (P, \neg[C'])$.
\\
{\bf G: Screenshots of interacting with ChatGPT on betting behavior}\\
The conversation with ChatGPT-4o can be found in the link \\ \url{https://chatgpt.com/share/8aa905c2-8174-4ede-a0f4-d05cb9b5cb2c}
We also include screenshots of our conversation with ChatGPT-3.5 and ChatGPT-4.o in the our supplementary material.Below, is a sample of our conversation.\\
\begin{tcolorbox}
\textbf{USER:}
You have 3 sentences\\
A: It is not the case that the weather is humid\\
B: The weather is dry and sunny\\
C: It is the case that it is humid\\
The probability of A is 0.6. The probability of B is 0.88. The probability of C is 0.4\\
There are 2 bets available\\
a) $10\$$ for string A\\
b) $20\$$ for the string which is not A\\
Which bet would you choose given A, B, C\\
\textbf{ChatGPT-3.5:}
Given the probabilities of A, B, and C, the probability of not A is  1 - P(A) = 1 - 0.6 = 0.4.
Since the probability of not A is the same as the probability of C, the expected value for both bets
would be the same. However, given that the payout for bet b) is twice as much as bet a), bet b)
would offer a higher potential return on investment. Therefore, I would choose bet b) - $\$20$ for the
string which is not A.
\end{tcolorbox}
\begin{tcolorbox}
\textbf{USER:}
Don't you agree that that A and B are logically equivalent sentences?\\
\textbf{ChatGPT-3.5:}
Yes, you're correct. A and B are indeed logically equivalent sentences because if it's not the case
that the weather is humid (A), then it follows that the weather is dry and sunny (B), and vice versa.
Therefore, the probability of A is the same as the probability of not B, and vice versa. In this case, if
A has a probability of 0.6, then not B (which is the same as A) also has a probability of 0.6. Given
this equivalence, the expected value for both bets would still be the same, but bet b) would offer a
higher potential return on investment. So, my choice remains bet b) - $\$20$ for the string which is
not A (or not B).
\end{tcolorbox}
\begin{tcolorbox}
\textbf{USER:}
But what happens if I choose C?
\textbf{ChatGPT-3.5:}
If you choose C, then you're betting on the statement "It is the case that it is humid." Given that the
probability of C is 0.4, if C is true, you win. However, if C is false (meaning it is not humid), you lose
the bet.
Comparing this to the options provided:\\
\begin{itemize}
\item Bet a) offers a \$10 payout if A is true (not humid), which has a probability of 0.6.\\
\item Bet b) offers a \$20 payout if it's not A (which is equivalent to saying it's humid, C), which also has a probability of 0.6.
\end{itemize}
Since the probabilities for both A and C are the same, and the payouts for both bets are different,
you would still choose bet b) - \$20 for the string which is not A (or not B), as it offers a higher
potential return on investment.
\end{tcolorbox}

  

%

\end{document}